\documentclass{article}

\pdfoutput=1 

\usepackage{amsthm,amsmath,amsfonts,amssymb}
\usepackage[authoryear]{natbib}
\usepackage[colorlinks,citecolor=blue,urlcolor=blue]{hyperref}
\usepackage{graphicx}
\usepackage{dsfont}
\usepackage{url}
\usepackage{color}

\usepackage{bm}
\usepackage{authblk}

\usepackage[margin=1.3in]{geometry}

\newcommand{\rcode}[1]{\normalfont\texttt{#1}}

\newcommand{\say}[1]{\lq\lq #1''}

\defcitealias{unece2011impact}{United Nations Economic Commission for Europe, 2011}
\defcitealias{un2022handbook}{United Nations, 2022}

\ifdefined\N                                                                % N, naturals
\renewcommand{\N}{\mathds{N}}                                                % N defined by "siunitx" (which we use), for "NEWTON"
\else
  \newcommand{\N}{\mathds{N}}
\fi
\newcommand{\R}{\mathds{R}}                                                 % R, reals
\ifdefined\C
  \renewcommand{\C}{\mathds{C}}                                             % C, complex
\else
  \newcommand{\C}{\mathds{C}}
\fi
\newcommand{\I}{\mathbb{I}}                                                 % I, indicator
\newcommand{\sumin}{\sum\limits_{i=1}^n}											% summation from i=1 to n
\newcommand{\xv}{\mathbf{x}}													% vector x (bold)
\newcommand{\yv}{\mathbf{y}}													% vector y (bold)
\renewcommand{\P}{\mathds{P}}                                               % P, probability
\newcommand{\E}{\mathds{E}}                                                 % E, expectation
\newcommand{\normal}{\mathcal{N}}                                           % N of the normal distribution
\newcommand{\Xspace}{\mathcal{X}}                                           % X, input space
\newcommand{\Yspace}{\mathcal{Y}}                                           % Y, output space
\newcommand{\Pxy}{\P_{xy}}                                                  % P_xy
\newcommand{\D}{\mathcal{D}}                                                      % D, data
\newcommand{\obs}[1][i]{\left(\xv^{(#1)},y^{(#1)}\right)}                                                                               % observation (x^(i), y^(i))
\newcommand{\Dset}{\left( \obs[1], \ldots, \obs[n]\right)}    % {(x1,y1)), ..., (xn,yn)}, data
\renewcommand{\xi}[1][i]{\mathbf{x}^{(#1)}}                                          % x^i, i-th observed value of x
\newcommand{\yi}[1][i]{y^{(#1)}}                                            % y^i, i-th observed value of y 
\newcommand{\Dtrain}{\mathcal{D}_{\text{train}}}                            % D_train, training set
\newcommand{\Dtest}{\mathcal{D}_{\text{test}}}                              % D_test, test set
\newcommand{\inducer}{\mathcal{I}}                                                % Inducer, inducing algorithm, learning algorithm 

\newcommand{\fx}{f(\mathbf{x})}                                                      % f(x), continuous prediction function
\newcommand{\fh}{\hat{f}}                                                   % f hat, estimated prediction function
\newcommand{\Lxy}{L\left(y, \fx\right)}                                               % L(y, f(x)), loss function
\newcommand{\ntest}{n_{\mathrm{test}}}                              % size of the test set
\newcommand{\ntrain}{n_{\mathrm{train}}}                            % size of the train set
\newcommand{\F}{\boldsymbol{F}}             % matrix of prediction scores
\newcommand{\Fi}[1][i]{\F^{(#1)}}             % i'th row vector of the prediction scores matrix
\newcommand{\lambdav}{\bm{\lambda}}											% lambda vector

\title{Evaluating machine learning models in non-standard settings: An overview and new findings}
\author[1,2,*]{Roman Hornung}
\author[3]{Malte Nalenz}
\author[3,2]{Lennart Schneider}
\author[3,2]{Andreas Bender}
\author[3,2]{Ludwig Bothmann}
\author[3,2]{Bernd Bischl}
\author[3]{Thomas Augustin}
\author[1,2]{Anne-Laure Boulesteix}
\affil[1]{Institute for Medical Information Processing, Biometry and Epidemiology, LMU Munich, Munich, Germany}
\affil[2]{Munich Center for Machine Learning (MCML), Munich, Germany}
\affil[3]{Department of Statistics, LMU Munich, Munich, Germany}
\affil[*]{Corresponding author: Roman Hornung, hornung@ibe.med.uni-muenchen.de}

\begin{document}

\maketitle

\begin{abstract}
Estimating the generalization error (GE) of machine learning models is fundamental, with resampling methods being the most common approach. However, in non-standard settings, particularly those where observations are not independently and identically distributed, resampling using simple random data divisions may lead to biased GE estimates. This paper strives to present well-grounded guidelines for GE estimation in various such non-standard settings: clustered data, spatial data, unequal sampling probabilities, concept drift, and hierarchically structured outcomes. Our overview combines well-established methodologies with other existing methods that, to our knowledge, have not been frequently considered in these particular settings. A unifying principle among these techniques is that the test data used in each iteration of the resampling procedure should reflect the new observations to which the model will be applied, while the training data should be representative of the entire data set used to obtain the final model. Beyond providing an overview, we address literature gaps by conducting simulation studies. These studies assess the necessity of using GE-estimation methods tailored to the respective setting. Our findings corroborate the concern that standard resampling methods often yield biased GE estimates in non-standard settings, underscoring the importance of tailored GE estimation.
\end{abstract}

\section{Introduction}

In supervised machine learning (ML) applications, the fitted models and their predictions on new data are of primary interest, but it is also generally crucial to quantify the expected predictive performance. While performance estimation of ML models is a well-discussed topic in the literature, most works are confined to simple data structures. They often rely on the \textit{i.i.d.\ }assumption, which posits that each data point is drawn independently from the same probability distribution. However, in practice, analysts frequently face non-standard settings, like spatial data. While the existing literature has highlighted certain concerns about performance estimation (for references, see the corresponding sections below) in this context, simulation studies in this paper aim to address gaps not previously covered. Standard resampling approaches can produce performance estimates that are either excessively variable or optimistically biased in non-standard settings. It is worth noting that such optimistic biases are generally more problematic than pessimistic ones, as outlined by \citep{Roberts:2017}. In typical resampling processes, based on the i.i.d.\ assumption, the observations are randomly allocated to any of the data subsets during the division into training and test sets. In many non-standard settings, however, more advanced resampling techniques may be required.

The main contribution of this paper is that it offers a unifying overview and guidance on using such techniques in various non-standard settings. While substantial literature exists on coping with most frequently encountered non-standard settings per se, for most of them, there is a shortage of results how to estimate performance as unbiasedly and accurately as possible. Hence, we complement, where appropriate, existing literature by our own insights derived from systematic simulation studies.

All non-standard settings considered in this paper are commonly found in official statistics, which makes it a prototypical field in need of the methods we discuss and evaluate. This importance is further underscored by the fact that a comprehensive quality assessment is constitutive in this field \citep{QF4SA:2022}. Therefore, the application examples included are influenced to some extent by this field. Nevertheless, the described methods and our simulation study results claim general validity across all areas where these settings occur.

Concretely, we consider the following five common non-standard settings: clustered data, spatial data, unequal sampling probabilities, concept drift, and hierarchically structured outcomes, which are described in detail in the subsections of Section~\ref{sec:main}. To better acquaint the reader with these settings and their practical manifestations, we provide here examples from official statistics for each:

\begin{itemize}
\item \emph{Clustered data} (cf.\ Section~\ref{subsec:clustered}): In the design of surveys and censuses, data are often clustered into groups, for example, households. Observations within these groups are typically more similar to each other than to observations from different groups.
\item \emph{Spatial data} (cf.\ Section~\ref{subsec:spatial}): Any type of data that includes geographic information can be considered spatial data. For instance, satellite imagery can be used to determine the type of land use in different areas (agricultural, urban, etc.). Here, regions close to each other are more similar in land use and other features than distant areas, leading to spatial correlations between neighbouring observations.
\item \emph{Unequal sampling probabilities} (cf.\ Section~\ref{subsec:nsrs}): In many surveys, not all units have the same probability of being sampled. For example, oversampling is a common practice to ensure that subgroups of particular interest (e.g., minority groups or companies with exceptionally high revenues) are adequately represented in the sample. As a result of using unequal sampling probabilities, the sample does not follow the same distribution as the population from which it is sampled.
\item \emph{Concept drift} (cf.\ Section~\ref{subsec:condrift}): The distribution of data arriving in streams can change continuously, or even  abruptly, over time. For instance, data on unemployment may be affected by steady changes in the skills required in the labour market or by economic structural breaks, respectively. 
\item \emph{Hierarchically structured outcomes} (cf.\ Section~\ref{subsec:hiercl}): In some classification problems, the observations do not belong to single classes but to a hierarchy of classes, with each class nested within a broader category. For instance, Class~1.3.2 may be a subclass within Class~1.3, which itself may be a subclass nested under the broader category of Class~1. Indeed, in official statistics, most classification systems are hierarchical. An example is the hierarchical occupational classification scheme ISCO-08 by the International Labour Organization (ILO) \citep{ISCO:2012}. The hierarchical structure of such schemes provides valuable information in statistical analyses, which has to be taken into account properly. 
\end{itemize}

Several empirical studies have conducted comparisons of commonly used resampling approaches under the assumption of i.i.d.\ situations \citep{kohavi.study.1995, BragaNeto:2004, Molinaro:2005}. For non-standard settings, however, literature on choosing appropriate approaches for performance estimation predominantly focuses on spatial data. We close this gap and conduct simulation studies comparing different approaches to performance evaluation when other types of non-standard settings are present. In these studies, we have deliberately structured the data-generating processes to be as consistent as possible across the different settings. For instance, all simulations maintain the same number of features. By ensuring similar structures for the data-generating processes, we minimize the potential for unintended result-dependent choices or overly specific configurations, thus enhancing the generalizability of our findings.

The remainder of this paper is structured as follows. Section~\ref{sec:general} introduces general terminology and concepts related to supervised learning and losses (Section~\ref{subsec:terminology}), performance metrics (Section~\ref{subsec:metrics}), generalization errors (Section~\ref{subsec:GE}) and data splitting and resampling techniques (Section~\ref{subsec:resampling}). Section~\ref{sec:main} discusses the different non-standard settings enumerated above. Depending on the specific setting, we provide varying amounts of background information, but always elaborate extensively on performance estimation. Please note that readers primarily interested in specific settings can refer to the corresponding subsections in Section~\ref{sec:main}, instead of reading the entire section. For ease of reading, we provide a summary of the principal conclusions at the end of each subsection. In Section~\ref{sec:discconcl} we address several topics. First, we describe how suitable performance estimation methods can be developed for non-standard settings not addressed in this paper. We then briefly address model selection and tuning parameter optimization for non-standard settings. Finally, we discuss potential limitations of the simulation studies.

\section{General strategies for evaluating ML models}
\label{sec:general}

In this section, we present a concise introduction to evaluating the performance of a supervised machine learning model, assuming the observations are i.i.d.\ samples.
We outline terminology, notation and fundamental concepts such as performance metrics, the generalization error and data splitting and resampling.
Readers familiar with these concepts can proceed to Section~\ref{sec:main}.
Notably, in the non-standard contexts covered in Section~\ref{sec:main}, where observations typically are not i.i.d., the resampling methods detailed in Section~\ref{subsec:resampling} are usually inapplicable. We describe them here because in Section~\ref{sec:main} they will be empirically compared with the resampling methods considered for the non-standard settings.

\subsection{Terminology}
\label{subsec:terminology}
The subsequent notation adheres to the conventions introduced in \citet{Bischl:2023}. Consider a labeled data set $\D = \Dset$ of $n$ observations, where each observation $(\xi, \yi)$ consists of a $p$-dimensional feature vector $\xi \in \Xspace$ and its corresponding label $\yi \in \Yspace$.
We make the assumption that $\D$ has been sampled i.i.d.\ from an unknown underlying distribution, denoted as $\D \sim (\Pxy)^n$. In the case of regression, the label space $\Yspace$ is defined as $\R$, whereas in classification, $\Yspace$ is a finite and categorical space with $\vert \Yspace\vert = g$ classes, where $\vert \cdot \vert$ represents the cardinality.
In supervised machine learning, an ML model is represented as a function $\fh: \Xspace \rightarrow \R^g$ ($g = 1$ for regression), mapping a feature vector from $\Xspace$ to a prediction in $\R^g$. In the case of classification, the outputs of $\fh$ in $\R^g$ are typically the predicted probabilities for the $g$ classes.

The objective of supervised machine learning is using a \emph{learner} $\inducer$ to train a model using a data set of $n$ observations sampled from $\Pxy$, aiming for good generalization performance on new, unseen observations generated by the same underlying data process. In general, the learner identifies $\fh$ by minimizing the \emph{empirical risk}, defined as the sum of the \emph{losses} across all observations $\sum_{i=1}^n L\left(\yi,f(\xi)\right)$. The \emph{loss function} $L : \Yspace \times \R^g \rightarrow \R^+_0$ measures the discrepancy between the predictions and the true labels.

A prevalent problem in ML is overfitting, where the model $\fh$ aligns too perfectly with the given data set $\D$, capturing artifacts of $\D$ that do not generalize to new data. For this reason, learners are configurable by hyperparameters $\lambdav$ that moderate the extent of adjustment of $\fh$ to $\D$. The value(s) of $\lambdav$ are not determined in the empirical risk minimization, but instead in a tuning step---typically via resampling procedures like cross-validation (CV) (see Section~\ref{subsec:resampling}). After this step, the empirical risk minimization takes place with the constraint that $\lambdav$ remains fixed to the value(s) determined during tuning. We will not delve into detail about hyperparameter optimization and the concept of nested resampling, for which the reader is referred to \citet{Bischl:2023}.

\subsection{Performance metrics}
\label{subsec:metrics}
After training an ML model $\fh$, a critical consideration is how to quantitatively measure its predictive performance. We seek to use a high-quality (i.e., consistent, efficient and ideally asymptotically normal) statistical estimator, which numerically quantifies the performance of our model when it is used to predict the labels of new observations drawn from the same data-generating process $\Pxy$ of the training data.

A general \emph{performance measure} $\rho$ for an arbitrary test set of size $m$ is defined as a two-argument function that maps the $m$-size vector of true labels $\yv$ and the $m \times g$ matrix of prediction scores $\F$ to a scalar performance value. This general definition based on sets is crucial for certain performance measures, such as the area under the ROC curve (AUC), and particularly for most measures used in survival analysis.
These measures are distinct in that they require consideration of the relative order or dependencies between multiple observations, as opposed to measures like the mean squared error (MSE), which also aggregate over multiple observations but treat each observation independently. 

It is possible to use the loss function applied during empirical risk minimization in performance measuring. However, it is common practice to select a performance measure depending on the specific prediction task at hand. During empirical risk minimization, this metric is often approximated using a computationally more efficient and potentially differentiable version of the performance measure.

\subsection{Generalization error} 
\label{subsec:GE}  
To mitigate the risk of overfitting, it is essential to assess the performance of every model using unseen test data, thereby ensuring unbiased estimation of its performance. We use $\Dtrain$ and $\Dtest$ to symbolize specified training and test sets with sizes of $\ntrain$ and $\ntest$, respectively. The \emph{generalization error} (GE) of a learner $\inducer$, trained on $\ntrain$ observations, is defined as the expectation of a performance measure $\rho$, with the expectation taken over both $\Dtrain$ and $\Dtest$. Performance measures that rely on non-point-wise losses require $\ntest\rightarrow\infty$ additionally.

The generalization error corresponds to the expectation of $\rho$ over infinitely many models, each fitted to a different realization of $\Dtrain$ of size $\ntrain$. This quantity tells us how well the learner (e.g., random forests) performs on average when trained and evaluated on data from the same distribution $\Pxy$. However, practitioners are generally interested in how well a fixed model $\fh$ performs on unseen data from the same distribution as that of the training data used to construct $\fh$. This type of performance can be measured by a quantity we refer to as the \emph{prediction error} (PE), obtained by holding $\Dtrain$ constant and taking the expectation solely over $\Dtest$.

The focus of this paper is on GE rather than PE due to technical considerations. Resampling techniques, such as CV (refer to Section~\ref{subsec:resampling}), are regarded as more suitable estimators of the GE than of the PE \citep{Bates:2023} because resampling generates a distinct model in each iteration. However, our results are also relevant for the PE because the latter can be expected to be similar to the GE. This becomes clear when acknowledging that the PE's expectation (taken over $\Dtrain$) is the GE.

\subsection{Data splitting and resampling} 
\label{subsec:resampling}
Usually, the GE must be estimated from a single given data set $\D$ of size $n$. The \emph{holdout estimator} serves as a straightforward approach to this task. First, $\D$ is randomly split once into a training set $\Dtrain$ of size $\ntrain$ and a test set $\Dtest$ of size $\ntest$. Second, the learner $\inducer$ is applied to $\Dtrain$ to produce a model $\fh$, which is then used on $\Dtest$ to obtain predictions that are assessed using the performance measure $\rho$.

The trade-off associated with the holdout estimator can be characterized as follows within the context of estimating GE:
(i) Due to the necessity of $\ntrain$ being smaller than the total number of observations $n$, splitting induces a pessimistic bias with respect to GE, as it does not utilize the entirety of the available data for training. In essence, the estimation is performed with respect to an incorrect training set size.
(ii) When the training set $\Dtrain$ is large, the corresponding test set $\Dtest$ becomes small, leading to increased variance in the estimator.
This trade-off is influenced not only by the relative sizes of $\ntrain$ and $\ntest$, but also by the absolute number of observations. Both the learning error estimation based on $\ntrain$ observations and the test error estimation based on $\ntest$ observations exhibits a saturating effect for larger sample sizes.
Nevertheless, a common heuristic consists of selecting $\ntrain = \frac{2}{3} n$ \citep{kohavi.study.1995, dobbin2011optimally}.

Resampling methods provide a partial solution to address this dilemma. These methods involve iteratively partitioning the available data into $B$ training and test sets, applying the second step of the holdout estimator to each split, and subsequently aggregating the $B$ resulting performance values. The aggregation is often performed by taking the mean. By repeatedly averaging over multiple splits the variance in estimating the generalization error is reduced \citep{kohavi.study.1995, simon.resampling.2007}. Moreover, the pessimistic bias inherent in a simple holdout approach can be minimized and nearly eliminated by selecting training sets of sizes close to the total number of observations ($n$). In general, there is little reason to prefer the holdout estimator over a resampling method except for computational efficiency reasons \citep{Hawkins:2003,raschka.model.2020a}.

Among the various resampling methods, \emph{$k$-fold CV} is widely employed. This technique involves partitioning the data into $k$ subsets (or \say{\emph{folds}}) of approximately equal size, using each subset successively as a validation set while fitting a model on the remaining data. In the case of small data sets, it is advisable to repeat the CV process with multiple random partitions and average the resulting estimates to reduce variability. This approach is known as \emph{repeated $k$-fold CV}. We focus exclusively on CV, despite the existence of various other resampling methods described in various sources, such as the work of \citet{Hastie:2009}. This choice is primarily due to the popularity of CV and the fact that most resampling methods share a common operating principle---repeatedly partitioning the data set into training and test subsets.

The number of folds $k$ has to be selected by hand. Employing a \emph{leave-one-out} strategy by choosing $k = n$ (i.e., fitting a model on all but one observation and evaluating the model on the left out observation and repeating this process for all observations) is not optimal as it increases the variance of the estimator [\citeauthor{James:2013}, \citeyear{James:2013}, p.~179, \citeauthor{Hastie:2009}, \citeyear{Hastie:2009}, p.~242]. Contrarily, repeated CV with many folds (but fewer than $n$) and multiple repetitions often yields better results \citep{bengio.no.2004}.

It is important to note that performance values obtained from resampling splits, especially CV splits, are not statistically independent due to the overlapping training sets, making the variance of resampling-based GE estimates notoriously difficult to estimate \citep{bengio.no.2004}. Consequently, carrying out statistical tests based on resampling-based GE estimates and creating confidence intervals around these estimates become challenging tasks. There are, however, quite a number of approaches that aim to estimate the variance of resampling-based estimators of quantities similar to the GE; refer to \citet{Efron:1997}, \citet{Austern:2020}, \citet{Bayle:2020}, or \citet{Bates:2023} for examples. However, it is unclear which of these estimators work well in which situations. For this reason, in an ongoing work by some of the authors, we are presently assessing these variance estimators' effectiveness in building confidence intervals for the PE in various different contexts.

\subsection{Conclusions on general strategies for evaluating ML models} 
\label{subsec:generalconcl}
Supervised ML aims to use labeled data to derive a prediction function for an outcome label, based on input features, which exhibits good predictive performance on new, unseen data. The performance of such a prediction function or model is examined using suitable performance measures. The selection of an appropriate measure depends on the nature of the outcome under consideration and the objective of the prediction.

In the context of performance evaluation, typically, two types of errors are distinguished: the GE (generalization error) and the PE (prediction error). The GE assesses the effectiveness of the learner, while the PE evaluates the performance of a specific ML model derived from the available data set at hand. Although the PE often receives more attention from practitioners, our attention is directed toward the GE due to technical considerations. Nonetheless, it is important to underline the close relationship that exists between these two.

Resampling methods, which involve repeated divisions of the data set into training and test data, are commonly used for GE estimation. These repeated divisions result in a good bias-variance trade-off of the estimate. In this paper, we exclusively focus on the most prevalent resampling method, $k$-fold CV. If computationally feasible, the value of $k$ should be chosen as a large value, yet smaller than $n$. Finally, estimating the variance of resampling-based GE estimators is challenging and a topic of ongoing research.

Background on supervised learning and performance evaluation is for example given in \citet{Hastie:2009} and \citet{James:2013}. Overviews of widely used performance metrics are provided  for example in \citet{Bischl:2023} and \citet{japkowicz.evaluating.2011a} with the latter focusing on the classification setting.
\citet{raschka.model.2020a} provides a review of model evaluation, selection, and algorithm comparison techniques, providing recommendations for best practices in machine learning research and applications, including discussions on bootstrap methods, CV, statistical tests, and alternative approaches for small data sets.
Recommendations regarding the choice of a resampling method further can be found in \citet{bischl.resampling.2012} and \citet{Boulesteix:2008}.
Finally, details regarding the difference between the PE and GE can be found in \citet{Bates:2023} and readers interested in hyperparameter optimization and nested resampling are referred to \citet{Bischl:2023}.

\section{Evaluating ML models in non-standard settings}
\label{sec:main}

This section is structured into subsections, with each one dedicated to examining one of the five non-standard settings: clustered data, spatial data, unequal sampling probabilities, concept drift, and hierarchically structured outcomes. As mentioned in the introduction, we have carried out simulation studies for all these scenarios, except for spatial data. An abundance of empirical results for spatial data can already be found in the existing literature. The purpose of the studies presented in this section is to evaluate and compare different methods of GE estimation for each non-standard setting, with an aim to establish preliminary guidelines for conducting GE estimation in these situations. Given that the concepts of \say{concept drift} and \say{hierarchically structured outcomes} may not be familiar to many readers, we offer further background information for these settings, expanding beyond elements associated with GE estimation.

\subsection{Clustered data}
\label{subsec:clustered}

\subsubsection{Nature and types of clustered data}
In the context of clustered data, also sometimes denoted as \say{correlated data} in the literature, a {\it cluster} is defined as a set of observations that \say{belong together} and are thus expected to be more similar to each other than observations from different clusters. Here, similarity may refer to the label $y_i$ and/or to the vector of features $\xi$. For example, if observations are individuals, households may be considered as clusters, because individuals from the same household are expected to share some characteristics. The same holds, say, for students from the same school or employees from the same company.

In this section we denote as $(\mathbf{x}^{(mi)},y^{(mi)})$ the $i$th observation from the $m$th cluster and as $n_m$ the number of observations in cluster $m$, with $\sum_{m=1}^Mn_m=n$ (where $M$ is the total number of clusters). In clustered data, it is generally assumed---as outlined above verbally---that $(\mathbf{x}^{(mi)},y^{(mi)})$ is more similar to $(\mathbf{x}^{(mi')},y^{(mi')})$ (with $i'\neq i$ standing for the index of another observation from the same cluster) than to $(\mathbf{x}^{(m'i')},y^{(m'i')})$ (with $m'\neq m$ standing for the index of a different cluster).

As opposed to Section~\ref{subsec:spatial}, there is no spatial structure, which makes the formalisation of the problem easier: the entirety of the information on the dependence structure is contained in a vector specifying which cluster each observation belongs to. Two observations are either in the same cluster or not. A special case of clustered data is obtained when clusters consist of several measurements taken from the same unit (e.g., from the same individual) at different time points. In this paper, however, we will not further consider this special case and assume data are not repeatedly collected over time.

\subsubsection{GE estimation for clustered data} 
\label{subsec:cluster}

The first question to ask when addressing GE estimation through resampling is how the model will be used in practice. If the model will be used to predict $y$ for new observations from the clusters included in the training data set, standard resampling techniques such as those described in Section~\ref{subsec:resampling} will certainly be appropriate. Here, the observations are randomly allocated to the data subsets when splitting $\D$ into training and test sets or (in the case of $k$-fold-CV) when splitting into $k$ folds. Consequently, in the same resampling iteration, observations from the same clusters can be found in both training and test sets. However, the much more common case in practice, at least as far as small clusters such as households are concerned, is that the goal is to use the model to predict $y$ for observations from new clusters not represented in the data set $\D$ used to fit the model.

 In the remainder of this section, unless stated otherwise, we will assume this second case. Here, standard resampling techniques have been shown to be optimistically biased several times in the literature \citep{brenning2008estimating, Saeb:2017, Gholamiangonabadi:2020, Kunjan:2021, Tougui:2021}. To understand that, imagine the extreme case where all observations from a cluster are identical and that this holds for all clusters, that is, $\forall m\ \forall i\neq i'\in\{1,\dots,n_m\},\  (\mathbf{x}^{(mi)},y^{(mi)})=(\mathbf{x}^{(mi')},y^{(mi')})$. Of course, in this extreme case, one would rather collect the data at the level of clusters and not at the level of observations, but we imagine this situation for didactic purposes. Further imagine that the cluster structure is ignored when splitting the data into training and test sets. We may have observation $(\mathbf{x}^{(mi)},y^{(mi)})$ in the training set and observation $(\mathbf{x}^{(mi')},y^{(mi')})$ in the test set. 
In our extreme case, however, these observations are equal, which essentially implies that the same observation is both in the training and test sets. By ignoring the cluster structure, we thus violate the principle of separating training and test sets, which generally leads to over-optimistic performance estimations. The same mechanism is in principle at work in cases where observations from the same cluster are only similar rather than identical.

\begin{figure}
  \centering
\includegraphics[width = 0.7 \textwidth]{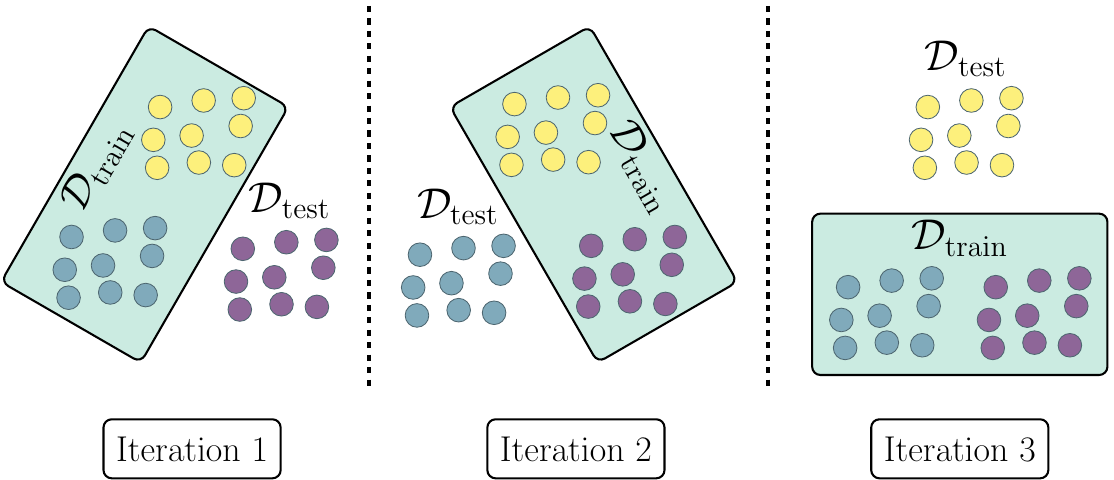}
\caption{Illustration of grouped CV, specifically showcasing the leave-one-object-out approach. Each dot represents an individual observation within a cluster, where observations from the same cluster are marked with the same color.}
\label{fig:groupedcrossval}
\end{figure}

To prevent over-optimistic performance estimations, all studies cited in the last paragraph recommend ensuring that in the resampling scheme the test observations originate from clusters absent in the training data. This can be implemented by performing resampling (e.g., CV) at the cluster (or subject) level---a method we will refer to as grouped resampling or grouped CV. This differs from the resampling at the observation level, which will henceforth be termed as standard resampling or standard CV. Grouped CV was originally suggested specifically for paired data ($n_m = 2$) by \citet{brenning2008estimating}. A commonly used special case of this method is leave-one-subject-out CV \citep{Gholamiangonabadi:2020, Kunjan:2021}---also known as leave-one-object-out CV [\citealt{Bischl:2024}, Chapter 3]. Here, each cluster is omitted once for evaluation, while training is conducted on the remaining clusters, see Figure~\ref{fig:groupedcrossval} for an illustration.

While most studies only consider grouped CV for GE estimation, \citet{Gholamiangonabadi:2020} also used it for model selection. They used it to choose between different alternatives in constructing a deep learning model. However, they did not explore whether standard CV would have led to a different optimal model. In a slightly different context, \citet{Bernau:2014} discovered that rankings of different ML learners can vary based on whether evaluations are performed on the same clusters or new ones. Instead of using grouped CV, \citet{Bernau:2014} employed cross-study validation, which involves training on individual clusters (or studies) and evaluating on all others. 

\citet{Saeb:2017} conducted a simulation study to examine how variability within and across clusters impacts the GE estimated with standard and grouped CV. Note that in this simulation study, as in \citet{Tougui:2021} or \citet{Kunjan:2021}, the clusters represented individual subjects, with each subject having consistent labels across their respective measurements. Here, as the variability across clusters increased, the error expected on new clusters also increased. However, standard CV did not reflect this trend in settings with small cluster numbers. \citet{Saeb:2017} also conducted a literature review in the field of smartphone- or sensor-based prediction of clinical outcomes. They found that more than half of the included studies used grouped CV, while the rest used standard CV. Grouped CV is also prevalent in medical research, as illustrated by \citet{pfau2020determinants} and \citet{kunzel2020determinants}. \citet{Saeb:2017} recommended the use of grouped CV not only for GE estimation but also for optimizing tuning parameter values. The choice of appropriate resampling methods for optimizing tuning parameters and model selection will be further discussed in Section~\ref{subsec:spatial} within the context of spatial prediction.

As previously mentioned, several studies have indicated that standard resampling techniques can be overly optimistic when applied to clustered data. However, many of these studies focused on specific scenarios or were based on individual data sets. For instance, \citet{Tougui:2021} and \citet{Kunjan:2021} considered cases where the clusters were represented by subjects with consistent labels across measurements. \citet{Gholamiangonabadi:2020} limited their focus to large clusters, and as highlighted earlier, \citet{brenning2008estimating} addressed paired data. In the upcoming section, we will present a simulation study that expands upon the existing literature discussed above. We will compare grouped with standard CV for cases with varying labels within clusters and varying sizes of clusters.

\subsubsection{Simulation study comparing GE estimation with and without respecting the clustering structure\newline}   

\paragraph{Objective and simulation model}
We conducted a small simulation study to compare the results of standard CV ignoring the clustering structure and grouped CV. For the simulation, we generated data from the linear model
\begin{equation}
y^{(mi)}=\sum\limits_{l=1}^5\beta_lx^{(mi)}_l+b_{m1}+b_{m2}x^{(mi)}_1+\epsilon^{(mi)},
\end{equation}
where $(\beta_1, \beta_2, \beta_3, \beta_4, \beta_5)^T = (1, 1, -1, 0, 0)^T$, $\epsilon^{(mi)}\sim\mathcal{N}(0,\sigma^2)$, $b_{m1}\sim \mathcal{N}(0,\sigma^2_1)$, and  $b_{m2}\sim\mathcal{N}(0,\sigma^2_2)$. The strength of the cluster structure is controlled by $\sigma^2_1$ and $\sigma^2_2$. Apart from the standard setting in which all $x_l$ values were sampled i.i.d.\ (from $\mathcal{N}(0,1)$), we considered two further settings, where the $x_1$ or $x_2$ values, respectively, were constant within clusters. Examples for such features would be family income or living space. We varied the number of observations $n_m$, the number of clusters $M$, and the strength of the signal through $\sigma^2$ (weak signal: $\sigma^2 = 1$ vs.\ strong signal: $\sigma^2 = 0.25$). Moreover, we considered settings with and without cluster-specific means and effects ($\sigma_s^2 = 1$ vs.\ $\sigma_s^2 = 0$, $s \in \{1,2\}$). Further details about the simulation design are described in Section~A.1 of the Supplementary Materials. 

\paragraph{Study design}
As a resampling technique, we used  (grouped) 5-fold CV, repeated ten times. For the standard variant, we ignored the information on cluster memberships, while for grouped CV the clusters rather than the single observations were randomly assigned to the folds. Lastly, we used linear models and random forests as learners and the MSE as error measure. In this simulation study, and in all other simulation studies presented in this paper, we conducted 100 repetitions per setting.

\begin{figure}
\centering
\includegraphics[width = \textwidth]{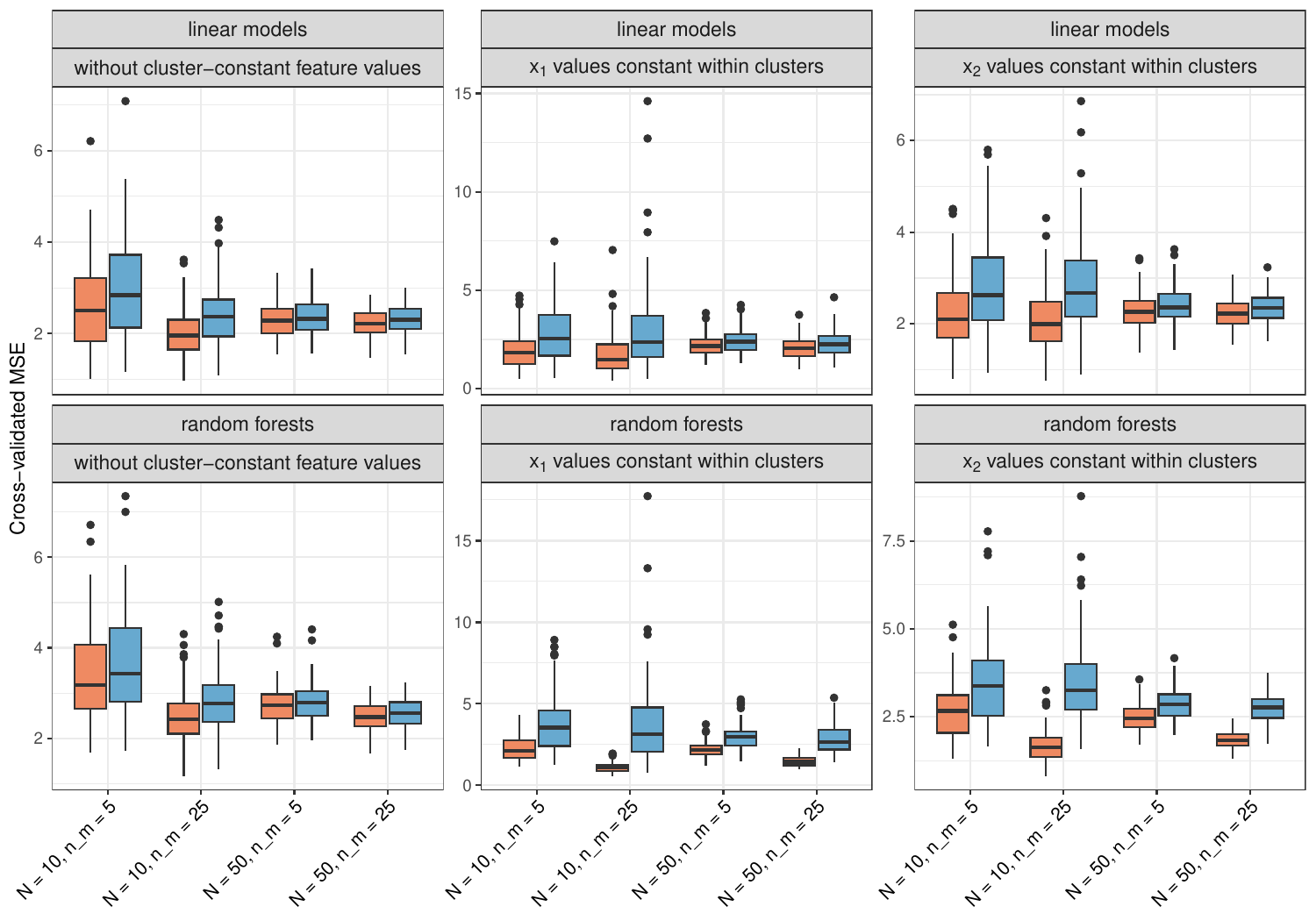}
\caption{Simulation on clustered data: cross-validated MSE values for each setting with cluster-specific means ($\sigma_1^2 = 1$), cluster-specific effects ($\sigma_2^2 = 1$), and strong signal ($\sigma^2 = 0.25$). The orange and the blue boxplots indicate the results obtained for standard and grouped CV, respectively.}
\label{fig:clustersim}
\end{figure}

\paragraph{Results}
Figure~\ref{fig:clustersim} visually represents the primary findings of our analysis on clustered data. For the standard setting in which all $x_l$ are sampled i.i.d., the MSE estimates obtained by standard CV are only slightly smaller than those obtained by grouped CV and only for small numbers of clusters ($N = 10$). For large numbers of clusters ($N = 50$), there are no notable differences between the results obtained for the two CV variants in the standard setting. In contrast, for the settings for which the $x_1$ or $x_2$ values, respectively, were constant within clusters, the MSE estimates obtained with standard CV are considerably smaller than those for grouped CV in many situations. Here, the differences are again smaller for larger numbers of clusters ($N=50$). 

Figure~\ref{fig:clustersim} includes results for all settings featuring cluster-specific means, cluster-specific effects, and a strong signal. These results are representative, as we observed no substantial differences in terms of the relative differences between the MSE estimates obtained for standard and grouped CV across settings that considered cluster-specific means, cluster-specific effects, or both. The signal's strength also did not notably impact the results in this respect.

All results are shown and described in Section~A.2 of the Supplementary Materials. The R code used to produce the results shown in the main paper and in the Supplementary Materials is available on GitHub (\url{https://github.com/RomanHornung/PPerfEstComplex}).

The findings from our study somewhat diverge from those reported in the existing literature. Past research consistently reported smaller error estimates for standard CV compared to grouped CV. In contrast, our simulations only mirrored these results for certain settings. Possible explanations for this discrepancy may be found in the characteristics of the data sets used in the literature, where the labels were constant within clusters, or there were only a few clusters, each with a considerable number of observations.

The observed strong underestimation of the error in standard CV can be reasonably expected when labels within clusters are constant. In such cases, ML models can more effectively predict the labels of observations from clusters that are already represented in the training data. Additionally, it is unsurprising to witness stronger underestimation by standard CV when larger clusters are present, as the ML learner has access to more information about each cluster, enabling it to fit the clusters better.

The simulation study of \citet{Saeb:2017} diverged from ours in several respects. Not only were the labels constant within clusters, but the features' distributions were also cluster-specific. The latter also can be expected to result in more homogeneous cluster, potentially amplifying the underestimation of error via standard CV.

\subsubsection{Conclusions for clustered data}
\label{sec:clustdataconcl}
When using clustered data for prediction modeling the cluster structure has to be taken into account during GE estimation. This can be performed by randomly assigning the observations to the folds on a cluster-by-cluster basis in CV, which prevents cluster overlap between training and test data. An exception is when the goal is to obtain predictions for new observations from clusters already existing in the training data. Here, standard CV should be used, that is, the observations rather than the clusters should be assigned randomly to the folds.

Our simulation study indicates that ignoring the cluster structure in GE estimation can lead to a slight, but non-negligible over-optimism. However, this effect can become stronger in the presence of features that take identical values within clusters, which is a situation frequently encountered in official statistics that is a focus of this paper. It has to be noted that the results from the existing literature consistently suggested strong over-optimism when the cluster structure is ignored. This pattern is likely attributable to the specific characteristics of the data sets and simulation designs used in these studies. Nonetheless, it underlines the importance of considering the cluster structure during GE estimation.

\subsection{Spatial data}
\label{subsec:spatial}

\subsubsection{Nature and types of spatial data}
\label{subsec:spatialtypes}
Spatial data exhibits spatial correlation due to its nature, which necessitates similar considerations with respect to resampling and GE estimation as with other correlated data types, such as clustered data (see Section~\ref{subsec:cluster}). Spatial data can be categorized into two types: discrete spatial data, collected, for example, at the level of federal states or other administrative units, and continuous spatial data, which includes the exact location of each observation, represented by coordinates $(j,k)$, for example, geolocated data with $j$, $k$ denoting longitude and latitude, respectively.

Discrete spatial data, in which observations are only known to be within a spatially defined region but without exact locations, can be considered a generalized case of clustered data. Observations in this case are grouped at state or district levels. However, such clusters exhibit not only within-cluster correlation but also spatial correlation, as data from neighboring administrative units could be more similar than data from distant units. Likewise, data collected at a finer resolution often demonstrates correlation in both features $\xi$ (e.g., temperature, infrastructure) and labels $\yi$ (e.g., unemployment rates).

\subsubsection{GE estimation through spatial CV}
\label{subsec:spatialapproaches}
To address the issue of spatial correlation, the literature typically recommends using resampling strategies that employ non-random allocation of observations to training and testing data sets, referred to as spatial resampling or \emph{spatial cross-validation} \citep{Schratz:2019}. As will be described below, there exist different variants of these strategies. All of them have in common that data used for performance evaluations are spatially separated from the training data to some extent, thereby reducing the correlation between the two. It has been frequently shown in the literature that failing to spatially separate training and testing data can lead to over-optimistic GE estimates; for examples, see \citet{Heikkinen:2012}, \citet{Wenger:2012}, \citet{Roberts:2017}, \citet{Schratz:2019}, and \citet{Schratz:2021}. This bias is particularly strong when prediction in new regions is a primary goal, as opposed to prediction in the space where the training observations are found, hereafter referred to as the \emph{observation space}.

Next, we give an outline of prevalent spatial CV approaches. Figure~\ref{fig:spatialcv} provides a visual representation of some of these methods, showcasing a single iteration of the spatial CV process in each case.  Details on how to set the parameters of these approaches and how to choose between them will be provided in Section~\ref{subsec:scvchoosing}. This synopsis is based on the descriptions provided by \citet{Schratz:2021}:

\begin{figure}
\centering
\includegraphics[width = \textwidth]{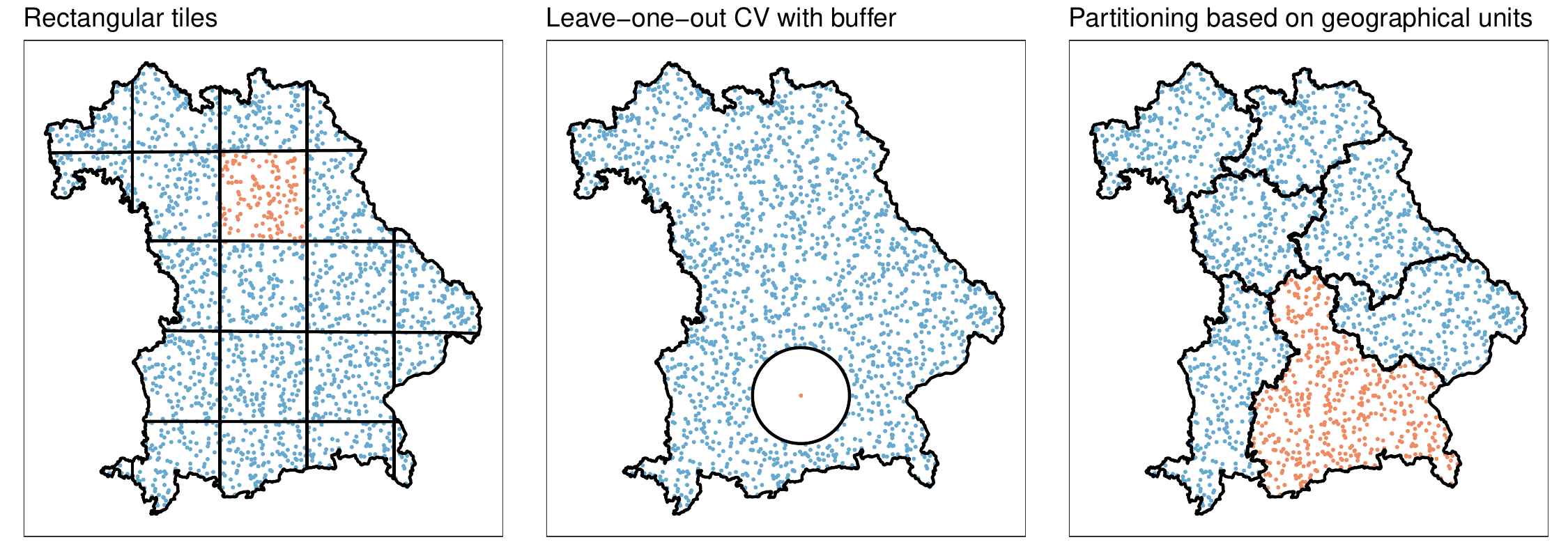}
\caption{Visualization of different spatial CV approaches. Each panel depicts a single CV iteration. The blue points represent the training data, while the orange points denote the test data. The spatial units in the last panel correspond to the seven administrative districts of Bavaria.}
\label{fig:spatialcv}
\end{figure}

\begin{itemize}
\item \emph{Single split into training and test data}: This is, arguably, the most basic method. Here, the data is spatially divided once into one training set $\Dtrain$ and one test data set $\Dtest$ in such a way that a single contiguous boundary between them can be established. It also permits the establishment of a buffer zone, a spatial region surrounding the boundary between $\Dtrain$ and $\Dtest$, with data within this zone excluded from both sets \citep{Heikkinen:2012}. As discussed in Section~\ref{subsec:scvchoosing}, this approach is suitable in situations in which the distance between the data intended for prediction and the labeled data is already known because it allows precise specification of where the test data lie relative to the training data. Despite not being a CV approach per se, it is still encompassed within the use of the term \say{spatial CV} in the following.
\item \emph{Rectangular tiles}: In this approach, the observation space is divided into uniform rectangular subspaces, termed \emph{blocks} \citep{Russ:2010}. The first variant of this approach uses a small number $k$ of blocks, each corresponding to a fold in a $k$-fold CV. The second variant involves a larger number of subspaces, which are randomly or systematically assigned to the $k$ folds in CV, akin to observations in a standard $k$-fold CV. This is referred to as CV \emph{at the level of the blocks}.
\item \emph{$k$ clustered groups}: This technique is similar to the rectangular tiles approach with $k$ blocks, but with the $k$ blocks determined through a clustering algorithm such as $k$-means clustering, rather than equally sized rectangles. Two variants exist: one with clustering based on the coordinates \citep{Russ:2010}, and the other with clustering based on the features \citep{valavi.blockcv.2018}, where for the latter the clustering is non-spatial (or not directly spatial).
\item \emph{Leave-one-out CV with buffer}: This is a spatial adaptation of the conventional leave-one-out CV. Unlike the latter, where all other observations except the test observation are used as training data in each iteration, a circular buffer zone is established around the test observation in each iteration in the spatial version. Observations within this buffer zone are excluded from the training data, spatially separating the test observation from the training data \citep{Rest:2014}. The choice of buffer zones in this and other methods will be further discussed in Section~\ref{subsec:scvchoosing}.
\item \emph{Leave-one-disc-out CV with optional buffer}: This approach is analogous to the previous one. However, in contrast, individual observations are not used for testing, but all observations within discs of a fixed diameter. The use of a buffer zone is optional in this method, unlike in leave-one-out CV with buffer. The number of test data sets is fixed to $k$, and the locations of the corresponding $k$ discs are randomly selected within the observation space \citep{Brenning:2005, brenning.spatial.2012}.
\item \emph{Partitioning based on geographical units}: In certain scenarios, for example, in the case of discrete spatial data (see Section~\ref{subsec:spatialtypes}), pre-existing spatial units may be available. These can be used as blocks in a spatial CV, which can also be done at the level of the blocks as in the rectangular tiles approach (see above).
\end{itemize}

All of the above approaches are available in the R package \say{mlr3spatiotempcv} \citep{Schratz:2022}, an extension package of the ML framework \say{mlr3}. The package is available online from the CRAN repository. It is important to note that the approaches described above are primarily applicable to continuous spatial data, as they necessitate the exact locations of the observations. Most of the literature presumes the presence of continuous spatial data, explaining the scarcity of resources about spatial CV methodologies for discrete spatial data. However, it should be highlighted that the same sort of spatial correlation exists in both continuous and discrete spatial data. Therefore, in the case of discrete data, we may simply use the centroids of each spatial region as substitutes for the precise locations and apply the approaches described above to these centroids.

\subsubsection{Choosing and configuring spatial CV approaches}
\label{subsec:scvchoosing}
The choice and precise configuration of a suitable (spatial) CV approach should be informed by the intended use of the model $f$ and the structure of the data. More specifically, the chosen spatial CV procedure should reflect the setting in which the ML model will be applied \citep{Schratz:2021}. For instance, if the objective is to fill in gaps in the observation space, the test data in the spatial CV should be closer to the training data than if the model is intended for application outside the observation space. Standard CV may suffice if predictions are to be made solely within the observation space \citep{Wenger:2012, Roberts:2017}. Predicting the prevalence of certain diseases is an example where both situations can occur. There may be applications where the goal is to fill in gaps within the observation space, as well as applications where the objective is to make predictions for new regions outside the observation space. For the former, rural areas could be underrepresented in health surveys, thus creating a need to extrapolate information from models trained on more urban areas. In the latter scenario, country A may have comprehensive disease prevalence data available, while country B may not. In this case, the aim would be to train an ML model using data from country A and obtain predictions for country B. 

Key parameters that influence the difference between training and test data folds in spatial CV are block width and buffer zone width. The larger the values of these parameters are, the larger the distances between the training and the test data sets are within spatial CV. Thus, the further the prediction area is from the observation space, the larger these parameters should be for realistic error estimates \citep{Roberts:2017}. The minimum width for the blocks and the buffer zone is often stipulated by the range parameter of the empirical correlogram, indicating the minimum distance between two points required for their labels to become uncorrelated \citep{Brenning:2005}. \citet{Roberts:2017} argue that larger widths may often be required due to structural features (such as temperature) causing label dependencies that extend beyond autocorrelation as measured by the correlogram. It is also crucial to measure autocorrelation in the raw data rather than in residuals from fitted models, to avoid underestimation of autocorrelation as fitted feature influences would have already accounted for some of it \citep{Roberts:2017}. Particularly when the aim is to identify influential features and measure their effect, it is important to choose large block or buffer zone widths. The identification of influential features with spatial data is generally challenging, with \citet{Wenger:2012} recommending feature selection based on plausible a priori hypotheses, as the features selected in this way are likely to have influence also in new areas outside the observation space. Spatial CV is recommended by \citet{Roberts:2017} even when the fitted models account for observation dependencies. 

Buffer zones should only be employed when predictions are intended for new areas, where the distance to these areas exceeds the distances between neighboring observations in the training data \citep{Schratz:2021}. Otherwise, spatial CV can result in GE overestimation because the sizes of the training data sets are reduced and extrapolations can occur. The latter means that feature values or combinations of feature values that are present in test data sets during spatial CV do not occur in the respective training data sets \citep{Roberts:2017}. If such a scenario does not arise in the application of the prediction model---for instance, because the distance between the data used for prediction later on and the training data is less than in the spatial CV---this could lead to an overestimation of the GE. Note, however, that as already mentioned in the introduction, in most cases overestimating the GE  is preferable to underestimating it \citep{Roberts:2017}.

Conversely, if the observation space is small, it may be difficult to generate sufficient independence between the training and test data through spatial CV because a sufficiently large buffer zone cannot be used in that situation \citep{Roberts:2017}. In such cases, obtaining a realistic error estimate from the available data may be challenging if the goal is to predict into new areas. This is because, without a sufficiently large buffer zone, the GE estimate will be overoptimistic for this goal. In these situations, \citet{Wenger:2012} recommend to keep the number of folds in spatial CV small. This leads to smaller training data sizes within CV and consequently, more conservative GE estimates. However, it can be challenging in practice to determine the optimal number of folds that yield a conservative bias sufficient to offset the over-optimism of the GE estimates. In contrast, \citet{Roberts:2017} suggest as a general rule, if computationally feasible, to keep the inherent bias of CV low, training data set sizes should be maximized by including only one block in each fold during spatial CV, instead of performing the latter at the level of the blocks (see the description of the rectangular tiles approach in Section~\ref{subsec:spatialapproaches}). If deciding a specific $k$ value in spatial CV is challenging, conducting a sensitivity analysis with varying $k$ values can be beneficial \citep{Wenger:2012}. 

Standard CV is typically repeated to reduce variability in the obtained error estimates. However, this is not usually feasible or meaningful with spatial CV, given the non-random division of data into training and test data sets that is determined by the spatial structure of the data \citep{Schratz:2021}. An exception might be $k$ clustered groups approaches, where, for example, $k$-means clustering depends on random initial values. Nevertheless, \citet{Schratz:2021} found that clusters can be nearly identical with different initial values used. In general, it is noteworthy that while spatial CV provides an indication of the model's expected performance, the actual performance can vary \citep{Wenger:2012}. Yet, spatial CV is preferable as it generally offers more realistic error estimates than standard CV \citep{Heikkinen:2012, Wenger:2012, Roberts:2017, Schratz:2019, Schratz:2021}.

Above, we discussed aspects of configuring appropriate approaches. Going forward, we will focus on selecting the most suitable one for the application at hand, based on the characteristics of the various options available. As previously mentioned, if the data are partitioned into pre-existing spatial units, these can be directly used as folds in spatial CV by employing the partitioning based on geographical units approach \citep{Schratz:2021}. 

If the data are irregularly distributed in the observation space, the rectangular tiles approach can result in significant variation in sample sizes across folds \citep{Roberts:2017}. In such cases, the $k$ clustered groups approach based on geographic coordinates is advisable, as here the areas associated with the folds align with the data distribution and the numbers of observations within the folds are comparable \citep{Schratz:2021}. 
The $k$ clustered groups approach based on the features should be used if the model will be applied to data containing feature values or feature value combinations not present in the training data \citep{Schratz:2021}. According to \citet{Roberts:2017}, although this yields less over-optimistic error estimates than standard CV, results can still be over-optimistic if spatial correlations are not considered. 

Spatial leave-one-out CV with buffer, similar to standard leave-one-out CV, has the downside of being highly computationally intensive, as each observation is used once as a test observation and the model must be refitted based on the corresponding training data in each of these cases \citep{Schratz:2021}. The leave-one-disc-out CV approach provides a less computationally demanding alternative. 

One issue with the folds in spatial CV is that it is not possible to be performed in such a way that each fold has a similar distance to the respective training data as the area intended for prediction. Thus, examining error estimates on specific folds might be more insightful if the distance between the data intended for prediction and the training data is known \citep{Roberts:2017}. We believe that in such scenarios, employing a single split into training and test data might be superior, as this allows for the specification of the test data's distance from the training data, and thus matching it to the distance of the data intended for prediction. Table \ref{tab:scvchoosing} provides an overview of the recommendations described above for choosing a particular approach in different situations.

\citet{Roberts:2017} argue that ideally, the test data should always originate from an independent area. However, it is debatable whether this is always the best approach, as the chosen validation procedure should reflect the nature of the application. Thus, validation on independent areas would be recommended only if the model will actually be applied to completely independent areas later on.

\begin{table}[ht]
\renewcommand{\arraystretch}{2}
\centering
\caption{Recommended spatial CV approaches in different application scenarios.}
\vspace{5pt}
\begin{tabular}{p{0.45\columnwidth} p{0.45\columnwidth}}
  \hline
        Application scenario & Recommended spatial CV approach(es) \\ 
  \hline 
        pre-existing spatial units (e.g.\ discrete spatial data) & partitioning based on geographical units \\ 
        irregularly distributed observations (e.g.\ involving clusters) & $k$ clustered groups approach based on geographic coordinates (rectangular approach is {\em not} recommended) \\
        application to data containing feature values or feature value combinations not present in the training data & $k$ clustered groups approach based on the features (spatial correlations should also be considered to avoid over-optimism) \\
        prediction to new regions & spatial leave-one-out CV (high computational burden); leave-one-disc-out CV with buffer (low computational burden)\\
        known distance between the data intended for prediction and the training data & single split into training and test data / examining GE estimates on specific folds \\
  \hline
\end{tabular}
\label{tab:scvchoosing}
\end{table}

\subsubsection{Model selection and tuning parameter optimization in spatial data applications}
\label{subsec:scvmodelsel}
As an initial point, it should be noted that the topics treated in this subsection are also applicable to the other non-standard settings discussed in this paper, as will be elaborated in Section~\ref{sec:discconcl}. We have already briefly touched on model selection and tuning parameter optimization in Section~\ref{subsec:cluster}, specifically in the context of the non-standard setting \say{clustered data}. Given the notable number of results available on them for \say{spatial data}, we have decided to dedicate an entire subsection to these topics within this specific context.

Traditionally, model selection for (generalized) linear models with i.i.d.\ data is performed using criteria like the AIC or the BIC. However, these are typically not applicable in ML because the models in this domain are usually not obtained by maximum likelihood estimation. In addition, the structure of spatial data leads to biased AIC and BIC values \citep{Roberts:2017} and overfitting to the observation space \citep{Wenger:2012}. When the goal is prediction within the observation space, CV may be appropriate for model selection. However, as pointed out earlier, even in this situation spatial CV is often recommended for GE estimation. More specifically, spatial CV should be used in this scenario if there are larger gaps between observations in the observation space, leading to cases for which the predicted observations have a distribution unrepresented in the training data even when predicting within the observation space.

The crucial role of choosing an appropriate validation method in spatial prediction, particularly for model selection, where the aim is to find the best-performing type of model (as opposed to, e.g., tuning parameter optimization discussed in the next paragraph), is supported by several empirical studies. These studies have suggested that different types of models can perform differently in predicting within and outside the observation space \citep{Brenning:2005, Heikkinen:2012, Wenger:2012, Schratz:2019}. These studies, however, offer some conflicting insights on the performance of various methods outside the observation space. This divergence is not unexpected, considering each study was based only on a single data set. For instance, \citet{Brenning:2005}, \citet{Heikkinen:2012}, and \citet{Wenger:2012} reported a more pronounced decline in prediction performance outside the observation space for tree-based methods, including random forests, compared to other methods. In contrast, \citet{Schratz:2019} found that random forests performed best in prediction both within and outside the observation space among the methods studied. Furthermore, while artificial neural networks showed good prediction performance outside the observation space in \citet{Heikkinen:2012}, according to \citet{Wenger:2012}  they underperformed in this scenario. Interestingly, both \citet{Heikkinen:2012} and \citet{Wenger:2012} observed that simpler model types, like linear models, generally tend to predict better outside the observation space compared to more flexible models. \citet{Schratz:2019} explain that very flexible models can adapt too closely to the specific conditions of the observation space, leading to decreased performance when predicting outside this space. Despite these differences, there is a positive correlation between the methods' performance inside and outside the observation space \citep{Heikkinen:2012}. This could explain why, despite typically being associated with a strong decline in performance outside the observation space \citep{Brenning:2005, Heikkinen:2012, Wenger:2012}, random forests also performed best outside the observation space in \citet{Schratz:2019}. 

The selection of tuning parameters $\lambdav$ in ML models dictates the degree to which these models adapt to the given data, thereby controlling model flexibility. Since the studies of \citet{Heikkinen:2012} and \citet{Wenger:2012} suggested that simpler models tend to perform better outside the observation space than more flexible models, it is logical to suggest that different tuning parameters could yield optimal performance depending on whether a model is applied within or outside the observation space. Typically, tuning parameter values are optimized by selecting those associated with the lowest CV error. To identify tuning parameter values that optimize model performance outside the observation space, spatial instead of standard CV can be used.

\citet{Schratz:2019} explored this possibility based on an extensive analysis using a data set on forest disease in Spain. Here, the difference between the predictive performance (outside the observation space) of models tuned with spatial and standard CV was minor for most model types, despite notable differences in the optimized tuning parameter values between the two approaches. Nevertheless, \citet{Schratz:2019} conceded the need for further validation using other data sets and model types. Based on an extensive empirical study, \citet{Ellenbach:2021} demonstrated that using external data with a different distribution from the training data for tuning often results in better performing models when applied to other data with a distribution different from the training data. This setting is consistent with that of spatial models used for prediction outside the observation space. Therefore, the results of \citet{Ellenbach:2021} suggest that, notwithstanding \citet{Schratz:2019}'s findings, using spatial rather than standard CV for optimizing tuning parameters could potentially result in improved predictions. Regardless, for the sake of consistency, \citet{Schratz:2019} recommend always using spatial CV to optimize tuning parameters if spatial CV is also employed for error estimation.

\subsubsection{Conclusions for spatial data}
\label{sec:spatialdataconcl}
In the GE estimation of spatial prediction models, it is most often critical to ensure a suitable spatial separation between training and test data. This is achieved by various procedures, collectively referred to as spatial CV. The choice and precise configuration of these procedures hinge on the data structure and the specific application scenario of the ML model. The primary determinant here is whether the model is intended for use within or beyond the observation space. In the former situation, and if, in addition, the training data are uniformly distributed in the observation space, standard CV without spatial separation might suffice.

Spatial CV procedures involve certain parameters, which govern the degree to which training and test data are separated. These procedures are more suitable for some types of applications than others, thus making the selection of an appropriate spatial CV method important. 

When used for model selection, the choice of spatial CV method also impacts which learner is selected. Existing research indicates that simpler learners tend to be more effective when aiming for robust predictions beyond the observation space. For consistency, spatial CV should also be used in tuning parameter optimization when used for GE estimation.

\subsection{Unequal sampling probabilities}
\label{subsec:nsrs}

\subsubsection{Reasons for and examples of unequal sampling probabilities}

In official statistics and other fields such as ecology \citep{schreuder.what.2001}, for both practical as well as principled reasons, sampling techniques that deviate from i.i.d.\ sampling are often used. Such designs are hereafter referred to as \textit{non-simple random sampling (NSRS)}, while i.i.d.\ sampling corresponds---for infinite populations to be sampled from---to the so-called \textit{simple random sampling} (SRS). The question of how NSRS affects the estimation of the GE is, to the best of our knowledge, an unexplored but essential area of research, given its high prevalence in practice, for example, when working with survey data.

A common motivation for using NSRS is that these designs can be much more efficient in terms of cost. Consider, for instance, a study that wants to measure the reading ability of school children. It may be difficult to obtain a population list of all school children. In practice, it would be even more expensive to perform SRS because this would involve interviewing/testing children from numerous, potentially distant schools. A more practical approach would be first to obtain a primary sample of $K$ schools.  In the second stage, a sample is taken from each selected school $k = 1,\dots,K$.\footnote{Note that the second stage sample does not have to be SRS either.} This approach is typically referred to as \say{two-stage sampling} \citep{valliant.practical.2018}.

To handle non-standard sample designs, it is helpful to rely on the specific perspective of sampling theory with its own terminology and notation. In this vain, one considers explicitly an unknown finite population $\mathcal{S}=\{1,\ldots, N\}$ of units $i$ with the corresponding data points  $(\xi, \yi),$ $i = 1, \dots, N,$ and a probability measure $P$ on the subsets of $\mathcal{S}$ specifying the sample design that determines which subset of $\mathcal{S}$ is actually observed as a sample. For a concretely realized sample $s  \subset \mathcal{S}$ of size $n=|s|$, the probabilities of having observed specific (pairs of) units are of particular interest: for every $i,j=1,\ldots,N,$ $i\not=j$, one calls  $\pi^{(i)} = P(i \in s)$ the  \textit{inclusion probability} of unit $i$ and $\pi^{(ij)} = P(i,j \in s)$ the \textit{joint inclusion probability} of $i$ and $j$. For the NSRS designs considered in this subsection, the inclusion probabilities $\pi^{(i)}$ differ between units in the population, that is, $\exists\, i,j : \pi^{(i)} \neq \pi^{(j)}$.\footnote{Here we do not study the very special NSRS situations where the inclusion probabilities of the single units are still equal, but joint (or so-called higher order) inclusion probabilities are not.}

 In the school example from above, the inclusion probability for each child is given by $\pi^{(i)} = \pi^{*(k)} \pi^{(i|k)}$ \citep{haziza.construction.2017}, where $\pi^{*(k)}$ denotes the inclusion probability of school $k$ and $\pi^{(i|k)}$ the conditional probability of including unit $i$ in the sample given the inclusion of school $k$. It is worth noting that two-stage sampling introduces a cluster structure, as discussed in Section~\ref{subsec:clustered}, with the schools acting as the clusters. Consequently, based on the findings of Section~\ref{subsec:clustered}, it is advisable to conduct resampling at the school level, rather than at the unit level, in scenarios akin to the one outlined above.

Even when an SRS design is possible, it may, nevertheless,  be advantageous to assign higher inclusion probabilities to certain units to reduce the variance of a population parameter estimate.  Consider an example where we want to estimate the average number of employees in a population using SRS of $n=100$ companies. Even though the simple mean estimator is unbiased, it will have a large variance. To reduce this variance, a common approach is to assign higher inclusion probabilities to large companies based on a known auxiliary variable $U$  that is approximately proportional to the label $Y$ \citep{horvitz.generalization.1952, valliant.practical.2018}. This approach stabilizes the mean estimate because larger companies tend to have a greater impact on the average number of employees. Note that in this and the previous example, the fact that the observations are sampled with different probabilities must be taken into account during parameter estimation to avoid biased estimates. In our example, one could use reported revenue as $U$ and set $\pi^{(i)} = n u^{(i)}/\sum_{i=1}^Nu^{(i)}$. This approach is known as \emph{sampling proportional to size} (PPS). Detailed overviews of different sampling designs can be found in \citet{skinner.introduction.2017}, \citet{valliant.practical.2018}, and \citet{haziza.construction.2017}. 

\subsubsection{GE estimation under NSRS}
\label{subsec:nsrsgeest}
To study the estimation of the GE of ML models, a small detour proves helpful. We first consider the estimation of the population mean under NSRS. Subsequently, the discussed concepts will be used to derive an unbiased estimator for the GE under NSRS.

Unbiased estimates of the population mean of $Y$ can be obtained from the sample $s$ of size $n$ using the \emph{Horvitz-Thompson estimator} of the mean
\begin{equation}\label{231002-HT}
\hat{Y}_{HT} = \frac{1}{N}\sum_{i=1}^nw^{(i)}y^{(i)}\,,
\end{equation}
where $w^{(i)} = (\pi^{(i)})^{-1}$. The resulting estimates typically have a much lower variance compared to the unweighted average $\hat{Y}_{SRS}$ \citep{horvitz.generalization.1952}. An alternative that can be used even when the size of the population $N$ is unknown is the H\'{a}jek estimator \citep{Hajek1971}
\begin{equation}\label{231002-hajek}
\hat{Y}_{HJ} = \frac{\sum_{i=1}^nw^{(i)}y^{(i)}}{\sum_{i=1}^nw^{(i)}}\,,
\end{equation}
which has a small bias under finite sample sizes. Nevertheless, it is often preferred over $\hat{Y}_{HT}$ \citep{sarndal.pinverse.1980}, since, intuitively, it compensates for potential extremes in the $w^{(i)}$ values.

When given a sample obtained using NSRS, conventional estimates of the GE may be associated with increased bias and variance as the i.i.d.\ assumption is clearly violated. For example, for a given test set with a fixed model $f$ and predictions $\bf{F}$, it can easily be shown that under PPS sampling, the sampling bias for the MSE is
\begin{align}
    \E&(\widehat{MSE_{SRS}} - \widehat{MSE_{PPS}})\nonumber\\
		&\hspace{-0.7em}=  \E\Big[ \sum_{i \in s_{SRS}}\big(y^{(i)}-\Fi\big)^2/n - \sum_{i \in s_{PPS}}\big(y^{(i)}-\Fi\big)^2/n \Big] \\
    &\hspace{-0.7em}= \frac{1}{N}\sum_{i=1}^N\Big[\big(\pi^{(i)}_{SRS}- \pi^{(i)}_{PPS}\big)\big(y^{(i)}-\Fi\big)^2/n\Big] \label{230922-1}\\
    &\hspace{-0.7em}= \frac{1}{N}\sum_{i=1}^N\Big[\big(n/N- nu^{(i)}/\sum_{i=1}^Nu^{(i)}\big)\big(y^{(i)}-\Fi\big)^2/n\Big]\,. \label{theoBIAS}
\end{align}
In the above equations, the subscripts $SRS$ and $PPS$ are used to denote that the corresponding quantities are related to the scenarios of simple random sampling and sampling proportional to size, respectively. The expressions in \eqref{230922-1} and \eqref{theoBIAS} directly imply that the bias depends on the deviation of $\pi^{(i)}$ from SRS and the dependency structure between $U$ and the error. If the inclusion probabilities $\pi^{(i)}$ vary independently from the error, no systematic bias is to be expected, although the variance will increase. If, however, the $\pi^{(i)}$ are correlated with the error, bias is to be expected, which is, for instance, the case in our example, where it is plausible that the prediction error increases with company size. 

This issue has been recognized in \citet{holbrook.estimating.2020}, who recommend the usage of Horvitz-Thompson estimates for the GE in the case of NSRS. However, the arising bias and unbiased estimators have not been studied extensively.

When performing CV on a sample obtained with NSRS, instead of considering a fixed model $f$, the situation requires further consideration. The reason for this is that the model $\hat{f}$ learned on an NSRS sample will also be affected by sampling bias if the differences in the sampling probabilities are ignored. However, in this paper, we focus only on the bias of the error estimation. A review of sampling-consistent learning methods under NSRS can be found in \citet{breidt.modelassisted.2017a} and applications in \citet{toth.building.2011a} and \citet{dagdoug2021model}.

For point-wise losses $\Lxy$ (cf.\ Section~\ref{subsec:metrics}), an unbiased estimator can be obtained by transferring the considerations on mean estimation via  the Horvitz-Thompson estimator (\ref{231002-HT}) from above, leading to 
\begin{equation}
\hat{\rho}_{L, HT} (\yv, \F) = \frac{1}{N}\sum_{i=1}^n w^{(i)}L(\yi, \Fi)\,. \label{HTloss}
\end{equation}

\begin{proof}
The proof utilizes a classical argument handling Horvitz-Thompson-type estimators, introducing, for $i=1,\ldots,N,$ Bernoulli distributed random variables $I^{(i)}$ to denote whether or not observation $i$ is included in the sample. Then, in the finite-population setting, $y^{(i)}$ and $\Fi$ can be seen as non-random \citep{horvitz.generalization.1952}; the randomness solely enters via the inclusion into the sample expressed by the inclusion variables $I^{(i)}$. Then 
\begin{align}
    \E&\Big(\frac{1}{N}\sum_{i=1}^n w^{(i)}L(\yi, \Fi)\Big) \nonumber\\
		&= \E\Big(\frac{1}{N}\sum_{i=1}^N I^{(i)} w^{(i)}L(\yi, \Fi)\Big) \\
    &= \frac{1}{N}\sum_{i=1}^N \E(I^{(i)}) (\pi^{(i)})^{-1}L(\yi, \Fi) \\
    &= \frac{1}{N}\sum_{i=1}^N \pi^{(i)} (\pi^{(i)})^{-1}L(\yi, \Fi) \\
    &= \frac{1}{N}\sum_{i=1}^N L(\yi, \Fi).
\end{align}
\end{proof}
Therefore, if $\pi^{(i)}$ is known, \eqref{HTloss} is an effective way to obtain design-unbiased estimators when performing CV or other re-sampling techniques with a point-wise loss. When the inclusion probabilities vary substantially, relying on the corresponding counterpart of 
(\ref{231002-hajek}) could be a worthwhile alternative. 

\subsubsection{Simulation study comparing GE estimation with and without correcting for NSRS}
To analyze the impact of NSRS on GE estimation and demonstrate the effectiveness of the Horvitz-Thompson estimator for error correction, we applied the concepts in a small simulation study based on our general simulation setting. 

\paragraph{Simulation model}
We generated data from the following linear model:
\begin{equation}
\yi = 5 + \sum\limits_{l=1}^5\beta_lx^{(i)}_l + \epsilon^{(i)},
\end{equation}
where $(\beta_1, \beta_2, \beta_3, \beta_4, \beta_5)^T = (1, 1, 0, 0, 0)^T$, and $\epsilon^{(i)} \sim \mathcal{N}(0,1)$.  Additionally, we generated an auxiliary variable (approximately) proportional to the target variable used for PPS sampling via\footnote{In the rare event that a draw did not lead to a positive value it was repeated.} $u^{(i)} \sim \mathcal{N}(y^{(i)}, 1)$  and set $\pi^{(i)} = nu^{(i)}/\sum_{i=1}^Nu^{(i)}$. The features with effect, $x_1$ and $x_2$, were sampled from a $Gamma(0.1,0.1)$ distribution, while the features without effect, $x_3$, $x_4$, and $x_5$, were sampled from $\mathcal{N}(0,1)$. The Gamma distribution for $x_1$ and $x_2$ was chosen to produce skewed label distributions, which under PPS sampling will lead to $\pi_i$ very far from $\pi^{(i)} = n/N$ underlying SRS. In this situation, based on equation~\eqref{theoBIAS}, we expected a large bias.

\paragraph{Study design}
The population size was varied ($N \in \{10000, 50000, 100000\}$) and PPS samples of size $n = N/100$ were drawn from the population with probabilities $\pi^{(i)}, i = 1, \dots, N$. We performed 5-fold CV, repeated ten times, on each sample, with and without correcting for NSRS using the Horvitz-Thompson theorem from Equation~\eqref{HTloss}. The GEs were approximated on separate test sets of size $2 \cdot 10^5$, where, for training, the entire available training data was used. Linear models and random forests were used as learners, and the MSE as error measure. We also included a variant for each simulated data set where feature $x_2$ was excluded after simulating the label, resulting in misspecified models.

\begin{figure}
\centering
\includegraphics[width = \textwidth]{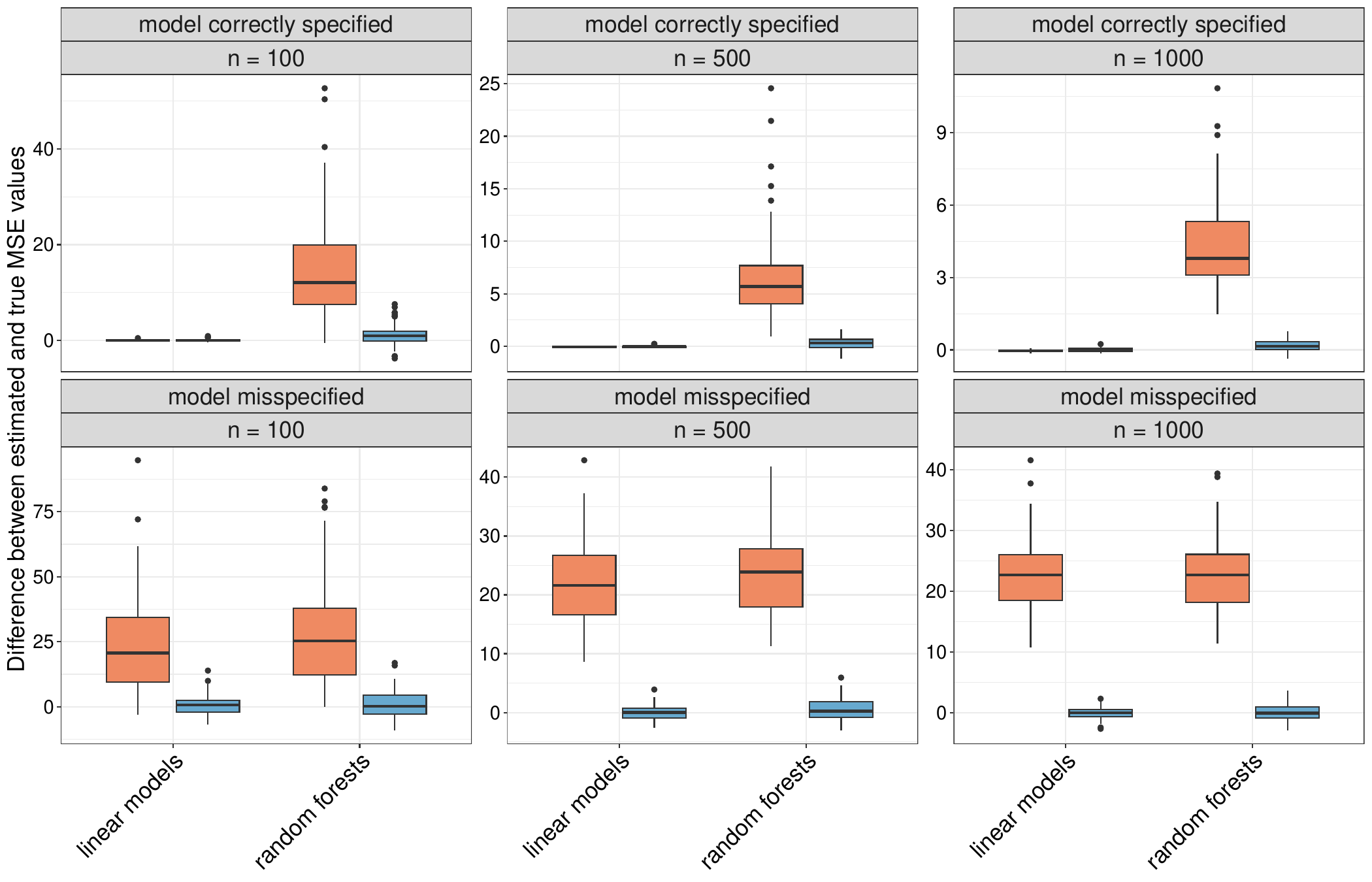}
\caption{Simulation on unequal sampling probabilities: differences between estimated and true MSE values. The orange and blue boxplots indicate the MSE estimates obtained using standard repeated CV and repeated CV with correction for unequal sampling probabilities via the Horvitz-Thompson theorem, respectively.}
\label{fig:nsrs}
\end{figure}

\paragraph{Results}
For the random forests, the GEs were overestimated when not corrected for NSRS. In contrast, for the linear models, the GEs were overestimated only when the models were misspecified. The GE estimates corrected for NSRS were unbiased across all simulation scenarios for both learners. The population size $N$ and sample size $n = N/100$ did not notably influence the bias observed in the GE estimates obtained without correcting for NSRS. The results are visualized in Figure~\ref{fig:nsrs}.

\paragraph{A potential interpretation of results}
The observed overestimation of the GEs for the random forests when not correcting for NSRS, even with correctly specified models, might be due to their flexibility. Random forests adapt locally to each region in the data distribution, which can lead to suboptimal performance in regions with few observations. The PPS sampling led to a higher number of observations with larger labels being sampled. However, since there were still fewer observations in this region compared to regions with smaller labels, the random forests likely performed poorly in this region. This could potentially explain the larger GE estimates for the random forests without correcting for NSRS. The (true) GEs were smaller, as the population had a greater proportion of observations with smaller label values compared to the training data. The random forests may have performed better for these smaller label values due to the higher density of observations in the region with smaller label values, both in training and test data. When correcting using the Horvitz-Thompson theorem, contributions from observations with larger labels are heavily down-weighted, which potentially led to the elimination of the upward bias in the GE estimates.

In contrast, linear models are less flexible but tend to be more robust to data sparsity in certain regions. When feature influences are linear, as in the simulation, linear models can be expected to provide accurate predictions even in areas with fewer observations. Therefore, it is conceivable that, unlike the random forests, the linear models did not perform worse in the region with larger label values; this would explain that the GE estimates for the linear models were not biased upward without correcting for NSRS. In the scenario with misspecified models, linear models exhibited similar behaviour to random forests, producing larger GE estimates without 
correction for NSRS. This likely occurred because, as with random forests, the prediction quality probably diminished for larger label values. A plausible reason behind this decline would be the increasing influence of $\beta_2x_2$ in this region, which is not considered in the misspecified models as $x_2$ is not used, leading to erroneous slopes.

\subsubsection{Conclusions for unequal sampling probabilities}
\label{sec:nsrsconcl}
There are instances in practice where observations are drawn from the population into samples with varying probabilities. As we demonstrated both analytically and through a simulation study, depending on the learner employed and the data distribution, if not accounted for, such unequal sampling probabilities can introduce bias into GE estimation. However, when the sampling probabilities are known, this bias can be corrected using the Horvitz-Thompson theorem, even if the ML model is misspecified.

\subsection{Concept drift}
\label{subsec:condrift}

The distribution of data arriving in streams can change over time, a phenomenon most commonly referred to as \say{concept drift} \citep{Gama:2014}. If not taken into account, this can lead to a decline in the predictive performance of ML models on future data. This is because the data-generating process $\Pxy$ that was present during training is no longer applicable to the new data obtained after training. If this is not taken into account when estimating the GE, it can lead to bias in the resulting estimates. We will observe this effect in the simulation study presented in Section~\ref{sec:condriftsim}. A further particular practical challenge in many situations is that it is not known how and when these changes in the data-generating process occur.

\subsubsection{Reasons for and types of concept drift}
\label{sec:concdrifttypes}

Naturally, the reasons for concept drift vary across different fields of application. \citet{Gama:2014} provides a comprehensive review of the occurrence of concept drift in various domains. In the field of official statistics, concept drift is a particularly common phenomenon as we will illustrate by five principled examples. One prevalent reason is changes in population structure; for instance, populations can age over time \citep{ec2023impact}. Second, social and economic changes, such as recent technological advancements in artificial intelligence, can alter relationships between features \citep{worldbank2019changing}. Third, environmental factors, including climate change or natural disasters, have widespread impacts on various aspects of society \citep{ipcc2022climate}, which can also contribute to concept drift. Fourth, changes in government policies or regulations, as seen recently with the COVID-19 pandemic, can lead to shifts as well \citep{worldbank2019changing}. Lastly, there can be modifications in the conduct of official statistics itself, such as changes in data collection methods, definitions, or classifications \citepalias{un2022handbook, unece2011impact}.

There are several forms of concept drift to distinguish:
\begin{itemize}
\item \emph{Pure feature drift}, also known as virtual drift \citep{Gama:2014}, occurs when only the feature distribution $\P_x$ changes over time, while $\P_{y|x}$, the conditional distribution of labels $\yi$ given the features $\xi$ remains fixed.
\item \emph{Pure label drift} \citep{Webb:2016} occurs when only the conditional distribution $\P_{y|x}$ changes over time, but the feature distribution $\P_x$ remains fixed.
\item \emph{Full concept drift} refers to situations where both $\P_x$ and $\P_{y|x}$ change over time (this definition slightly differs from that in \citet{Webb:2016}).
\end{itemize}
Pure feature drift is typically regarded as the least concerning type in the concept drift literature, because prediction models acquire knowledge on the conditional distribution $\P_{y|x}$ \citep{Gama:2014}. However, this assumes that the model is specified correctly and can successfully extrapolate to regions beyond the scope of the feature values observed in the training data. A prime example of such circumstances constitutes regression models that incorporate all influential features and linearly model their influences. Under these conditions, and importantly, if the true feature influences are also linear, the model can accurately represent $\P_{y|x}$ even outside the range of feature values observed in the training data. This trait of regression models significantly contributes to their widespread use in the medical field, where data sets frequently deviate from representing the general population due to numerous factors, such as specific inclusion and exclusion criteria for medical studies. For instance, studies might only involve patients below a certain age \citep{Vach:2012}. Applying a prediction model trained on these data to new data epitomizes pure feature drift, as the new data may include patients from age groups not represented in the training data. As demonstrated in the simulation study presented in Section~\ref{sec:condriftsim} using the example of random forests, ML models, despite their impressive adaptability within the range of feature values in the training data, may struggle to extrapolate, even if the feature influences are linear. However, in the presence of nonlinear feature influences, traditional regression models are also susceptible to poor performance when extrapolating to feature values outside the training data range. These models can still reasonably depict the dependence structure in the training data, even with moderate non-linearity. However, in the case of non-linearity beyond the scope of the feature values observed in the training data, the ability of linear extrapolation to accurately represent the true dependence structure decreases as the extrapolation extends further from the range of feature values observed in the training data.

\begin{figure}
\centering
\includegraphics[width = \textwidth]{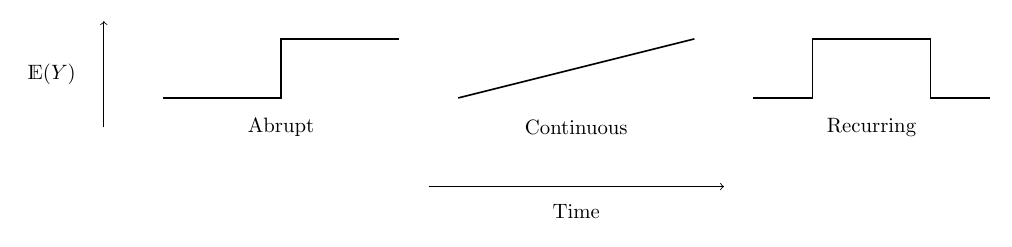}
\caption{Illustration of abrupt, continuous, and recurrent concept drifts. A drift in the marginal distribution of the label $Y$ over time is shown.}
\label{fig:concdrift}
\end{figure}

The above categorizations of concept drift are based on whether $\P_x$ or $\P_{y|x}$ change over time. Another way to classify concept drift is by examining how these distributions change over time:

\begin{itemize}
\item \emph{Abrupt}: The distribution remains fixed until a specific point in time and then suddenly changes.
\item \emph{Continuous}: The distribution changes continuously over time and potentially only within a specific time interval. As defined by \citet{Webb:2016}, the change is \say{incremental} if it progresses consistently in one direction, and \say{gradual} if it changes continuously over time without a consistent direction. \citet{Webb:2016} also distinguish a related concept drift type, for which the data initially follow distribution $A$, then transition to distribution $B$, and in between, there are periods when the data follow either distribution $A$ or $B$.
\item \emph{Recurring concepts}: The distribution changes at some point but later returns to its original form \citep{Gama:2014}.
\end{itemize}
Figure~\ref{fig:concdrift} provides an illustration of these three types of concept drifts.

In practice, concept drifts may not always occur in isolation; sequences of concept drifts can take place. For instance, several abrupt drifts may occur within a particular time range. Such sequences of concept drifts can be further characterized by their predictability and frequency \citep{Minku:2009}. An example of predictable drifts would be those that occur seasonally. Lastly, the examples of concept drift occurrences presented at the beginning of this subsection fit into different categories based on the categorizations outlined above. Understanding how each example aligns with these categories can deepen insights into the various dynamics of concept drift.

\subsubsection{Strategies for detecting concept drift and mitigating its impact}

As indicated above, when concept drift occurs, the predictive performance of ML models deteriorates on future data. This is because the distribution of the data to which the ML model is applied differs from the distribution of the training data. In the case of incremental drift, these two distributions will become more disparate as time progresses. The deteriorated predictive performance can lead to suboptimal decision-making and increased costs. Detecting and adapting to concept drift is therefore essential. Additionally, as pointed out at the start of this section, overlooking concept drift during the estimation of the GE can result in biased GE estimates. Section~\ref{sec:condrifteval} will delve  further into methodologies for estimating the GE considering concept drift, while Section~\ref{sec:condriftsim} will present the outcomes of a simulation study comparing some of these methods.

\paragraph{Detection}
Approaches for detecting concept drift can be based on the error rates of the ML models or on the data distribution $\Pxy$ itself \citep{Lu:2018}. The choice between these two depends on the application. If the primary interest is in prediction, it is advisable to use an approach based on the error rate rather than the data distribution, as the data distribution may change significantly without greatly affecting prediction performance (e.g., in the case of pure feature drift). An early and straightforward error rate-based approach, which performed best in a benchmark study by \citet{Goncalves:2014}, is the Drift Detection Method (DDM) \citep{Gama:2004}. This method is built on the idea that the prediction error should decrease with more training observations under stationarity but should increase as soon as concept drift emerges. When the error rate surpasses a \say{warning level}, the algorithm begins gathering new training observations while still using the existing model for prediction. Once the error rate exceeds a \say{drift level}, the collection of new training observations ceases, and the model is retrained using the recently acquired training data. Since the warning and drift levels are based on the Binomial distribution, assuming Bernoulli-distributed errors, the method in its basic form is only applicable to binary outcomes. Data distribution-based approaches, generally more computationally demanding, involve measuring the dissimilarity, such as using the total variation distance between the distribution of the historical data and that of the new data. These approaches necessitate fixed window sizes for both historical and new data, with the new data window moving forward until concept drift is detected \citep{Lu:2018}. Both error rate-based and data distribution-based approaches typically employ statistical tests. Initially, only single tests were used, but later tests using multiple hypotheses were introduced. The application of multiple hypothesis tests varies across these approaches. For instance, three-layer drift detection \citep{Zhang:2017} allows for separate testing of pure feature drift, pure label drift, and full concept drift.

\paragraph{Incremental and online learning}
One of the most straightforward ways to address concept drift is to retrain the model each time a batch of new observations arrives, using both the old and the new observations. This approach is known as incremental learning. Online learning is a special case of this, wherein the model is updated each time a new observation becomes available \citep{Zhang:2019}.

\paragraph{(Adaptive) windowing}
In data streaming scenarios with an infinite stream of data, it is not feasible to use the entire history for real-time training and prediction \citep{Zhang:2019}. In such cases, it is necessary to \say{forget} older observations and use only the latest observations for retraining the model. This process is called \emph{windowing}, where a time window is used to consider only observations within a specific time frame for training. Apart from computational considerations, another reason to use only recent data for training is that, under strong continuous concept drift, the distribution of old observations becomes too dissimilar from that of the current observations. Using only recent observations for training ensures that the resulting ML models can rapidly adapt to the current data distribution. Choosing the width of the time window is both important and challenging. If the window size is too small, the training data will have too few observations, resulting in unstable ML models. In contrast, if the window size is too large, the training data will include observations too disparate in distribution from the current observations. Some approaches, such as ADWIN \citep{Bifet:2007}, choose the window size adaptively in a data-driven manner \citep{Lu:2018}.

\paragraph{Model updates and ensemble methods}
In addition to completely retraining ML models, another option is to update existing models in response to concept drift. As this strategy involves less drastic changes to the model, it may be more suitable for cases where concept drift is less severe or only occurs in certain regions of the data distribution (e.g., for specific features). This strategy is not applicable to all ML models, but specialized approaches can achieve this. According to \citet{Lu:2018}, many such approaches are based on the decision tree algorithm, as trees can adapt flexibly to different regions of the feature space; examples include \citet{Hulten:2001}, \citet{Gama:2003}, \citet{Rutkowski:2014}, \citet{FriasBlanco:2016}, and \citet{Ducange:2021}. Another method using model updating in response to concept drift is DELM \citep{Xu:2017}, which is based on the Extreme Learning Method (ELM) \citep{Huang:2006}, a fast neural network using only a single layer. DELM adds new nodes to the network when the error rate increases, indicating the emergence of concept drift.

Ensemble methods can also be used to adapt to concept drift. These approaches fall into the same category as the methods discussed in the previous paragraph, although they are conventionally treated separately in the concept drift literature. Ensemble methods are suitable for handling concept drift due to the fact that they consist of separate models, or \say{base classifiers}, that capture different aspects of the data distribution, and new base classifiers can be added, removed, or re-weighted dynamically. As time progresses under concept drift, some aspects captured by the base classifiers  become more or less important, and new aspects arise. Consequently, some ensemble methods for concept drift remove or add base learners dynamically or alter the weighting of the base learners. There are also methods with online base learners that update the learners themselves in the presence of concept drift. Additionally, certain ensemble methods are specifically designed to deal with recurring concepts. Many of these ensemble methods use trees as base learners. Refer to the overview papers by \citet{Gama:2014} and \citet{Lu:2018} for more details on ensemble methods under concept drift and concept drift methodology in general.

The field of medical statistics also offers various methods designed to update models under concept drift, as described in \citet{Steyerberg:2019}. Many of these methods focus on parametric models, where individual parameters, such as the intercept, are re-estimated based on the newly available data. A common approach here is to shrink the parameters estimated on the newly available data towards those from prior to the model update. This technique helps mitigate the potential for high variance in parameter estimates derived from the new data, as the latter are typically limited in size.

\subsubsection{GE estimation under concept drift}
\label{sec:condrifteval}
The literature on performance evaluation for streaming data under concept drift is rather limited. One of the most widely adopted approaches for estimating the GE in streaming data is \emph{prequential (predictive sequential) validation} \citep{Dawid:1984}, which is often referred to as \say{time-series CV} \citep{Hyndman:2018}. With this method, the model is first trained using an initial portion of the data and then applied to the next observation. Afterward, this next observation is included in the training data, the model is updated or retrained, and evaluated on the subsequent observation. This process is iteratively repeated until all (except for the last) observations have been used for training. \citet{Gama:2013} found that the prequential approach, combined with sliding windows or fading factors that weigh recent observations more strongly, allows effective detection of concept drift as it can swiftly adapt to changes over time.

\emph{Out-of-sample validation}, a technique commonly employed in time series forecasting \citep{Bergmeir:2018}, can also be applied to streaming data under concept drift. It is similar to hold-out validation, with the distinction that the test set is located at the end of the observation period, resulting in a temporal separation between training and test sets. The literature on out-of-sample validation typically does not specify which data should be used for training and testing. We contend that this decision should be informed by the available data set size $n$ and the error estimation goal. When working with a large data set, only a small portion of the data at the end of the observation period should be designated as test data. This is because this portion is large enough in this situation, and the GE estimate is likely to be more realistic since the test data at the end of the observation period tends to be more similar to subsequent observations. Conversely, with a smaller data set, a larger portion of test data is needed to avoid excessive variance in the GE estimate due to the small test data set size. However, using a larger proportion of test data leads to upward bias in the calculated GE estimates. This happens for two reasons. First, out-of-sample validation is subject to a bias-variance trade-off, similar to that observed for the holdout estimator (discussed in Section~\ref{subsec:resampling}). Second, when the test data set is large, the last observations used for training become less similar to the last observations of the observation period. 

\begin{figure}
  \centering
\includegraphics[width = 0.7 \textwidth]{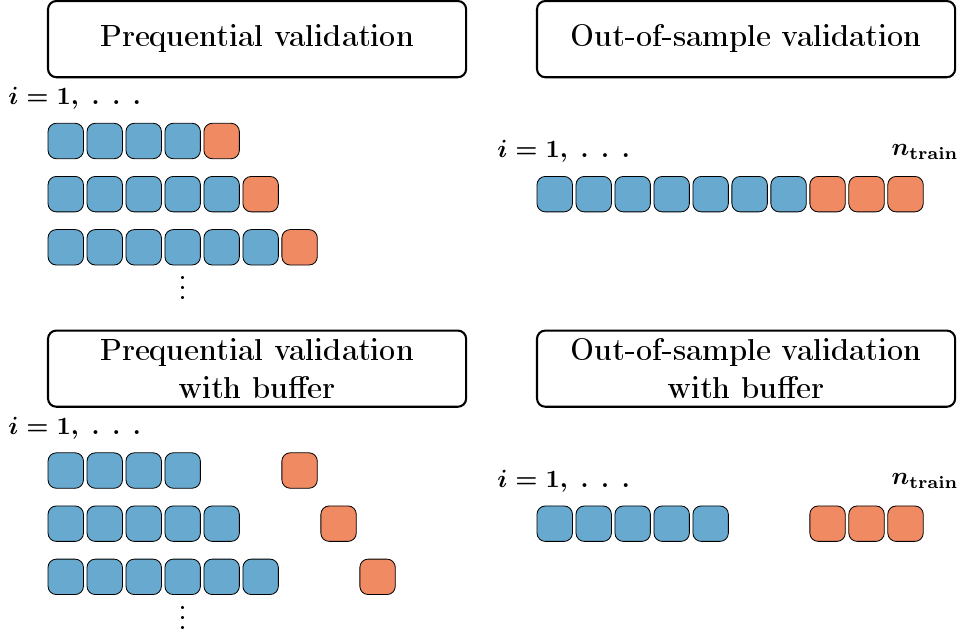}
\caption{Illustration of prequential validation and out-of-sample validation with and without a buffer zone. Each box represents a single observation, with blue points representing training data and orange points representing test data.}
\label{fig:concdriftgeest}
\end{figure}

If the error estimation's objective is to estimate the GE expected immediately after the observation period, the test data $\Dtest$ should directly follow the training data $\Dtrain$ without introducing a buffer zone. However, if the aim is to employ the ML model for an extended period and estimate the error expected further into the future, a buffer zone should be placed between the training and test data. The buffer zone's width should be selected based on the model's intended duration of use. Nevertheless, it is important not to attempt to estimate the GE too far into the future, as the data distribution may change unpredictably in the distant future. Figure~\ref{fig:concdriftgeest} provides a graphical representation of both prequential and out-of-sample validation, illustrating scenarios with and without the inclusion of a buffer zone. In the simulation study presented in the next subsection, we assess out-of-sample validation with and without a buffer zone.

The controlled permutation approach \citep{Zliobaite:2014} artificially induces concept drift in the data by iteratively permuting the order of the observations in specific ways. \citet{Zliobaite:2014} proposes three different permutation techniques, each designed to induce one of those three types of concept drift that refer to how the data distribution changes over time (see Section~\ref {sec:concdrifttypes}): abrupt changes, continuous changes, and recurring concepts. For instance, to induce abrupt changes, blocks of consecutive observations of random length are permuted, while to simulate recurring concepts, randomly chosen observations are relocated to the end of the observation period, keeping their original order. This results in recurring concepts because the sequence of distribution changes in the original stream are repeated twice in the permuted stream. However, these techniques do not seem to have gained widespread acceptance. \citet{Zliobaite:2014} justifies them by arguing that the results from the classical prequential approach may not be highly generalizable, as they are based solely on the concept drifts occurring in the respective training data. Nonetheless, in the case of controlled permutation, it seems challenging to determine whether the artificially induced concept drift patterns and the strengths of the associated concept drifts are realistic concerning the specific applications. This problem does not arise with the prequential approach, as it uses changes that have actually occurred.

\subsubsection{Simulation study comparing different approaches to GE estimation under concept drift\newline}
\label{sec:condriftsim}

\paragraph{Objective and scope}
This simulation study assesses the performance of (non time-series) CV, time-series CV (i.e., prequential validation), and out-of-sample validation in the context of data with concept drift, in order to determine which method is generally more favorable. It was expected that CV would exhibit lower variance but strong optimistic bias for the GE expected after the observation period (at different prediction horizons), as it does not account for the temporal structure. Conversely, out-of-sample validation was anticipated to exhibit larger variance (as it uses fewer observations for testing) but lower bias, as it does consider the temporal structure. Time-series CV can be seen as an intermediate solution, see below for more details on the simulation design. 

Note that in streaming applications with very large numbers of observations, using CV for performance evaluation is likely not meaningful, as the potential advantage of CV, having a smaller variance, is not present, since any meaningful evaluation procedure can use large numbers of observations for testing if there are very large numbers of observations available. Accordingly, \citet{Gama:2013}, who considered streaming applications with potentially infinite numbers of observations, argued that CV should only be used in applications where the data distribution does not change with time. In contrast, we consider applications with smaller numbers of observations $n$, where the comparatively small variance of CV could be advantageous. To our knowledge, no prior research has investigated how well different approaches perform in estimating the future GE under concept drift; our simulation aims to fill this gap.

\paragraph{Simulation model}
We simulated data according to the following model: 
\begin{equation}
y^{(t)}=\beta_0(t) + \sum\limits_{l=1}^5\beta_lx_l^{(t)}+\epsilon^{(t)}, \quad t \in [0,1],
\label{eq:conddrift}
\end{equation}
where $(\beta_1, \beta_2, \beta_3, \beta_4, \beta_5)^T$ $=$ $(2, -1, 2, 0, 0)^T$, $x_{l_1}^{(t)}\sim\normal(\mu_{l_1}(t), 1)$ for $l_1 \in \{1,2,3\}$,  $x_{l_2}^{(t)}\sim\normal(0, 1)$ for $l_2 \in \{4,5\}$, and $\epsilon^{(t)}\sim\normal(0,\sigma^2(t))$. The time-dependent label mean $\beta_0(t)$ and variance $\sigma^2(t)$ of $y^{(t)}$ as well as the time-dependent feature means $\mu_1(t)$, $\mu_2(t)$, and $\mu_3(t)$ depended linearly on $t$. This corresponds to a constant degree of change in the distributions of $y^{(t)}$ and $x_{l_1}^{(t)}$ ($l_1 \in \{1,2,3\}$), meaning that we simulated an incremental drift (see Section~\ref{sec:concdrifttypes}). 

We considered different strengths of feature and label drift. Moreover, we divided the period under consideration into ten equally wide intervals, referred to as \say{seasons}. The first eight seasons were considered the collection period for the training data, hereafter called the training period, and the last two were considered future seasons for which predictions should be obtained. In Section~B.1 of the Supplementary Materials, we describe the study design, including the simulation design, in full detail.

Note that, in practice, periods of concept drift are often preceded and followed by periods with constant data distributions. However, we deliberately simulated a drift throughout the entire period because we wanted to investigate how well the compared evaluation methods performed in estimating the GE one and two seasons ahead. If the concept drift had not continued until the last season, the GE in the ninth and tenth seasons would have been the same.

\paragraph{Evaluation methods and study design}
We used 8-fold CV, repeated ten times, 8-fold time-series CV (prequential validation), and out-of-sample validation as evaluation methods. For the time-series CV, the folds coincided with the seasons, and unlike with standard prequential validation, we did not re-learn the prediction rules after each new observation but after each season. This was motivated by practical reasons, as otherwise the calculations would have taken a long time. However, it also reflects the practical situation in official statistics, where ML and statistical models are often applied over longer periods without updating or retraining them \citep{Meertens:2022}. For out-of-sample validation, we used the first seven seasons as training data and the eighth season, that is, the last season of the observation period, as test data.

We also considered variants of the time-series CV and the out-of-sample validation, where we placed a buffer zone between the training and test data, similar to the spatial validation in Section~\ref{subsec:spatial}. More precisely, we did not use the next season but the season after next as test data. As a result, for the out-of-sample validation, we used the first six seasons as training data and the eighth season as test data. The resulting error estimators of these variants should estimate the GE in the season after next. Accordingly, we also approximated the GE at different points in time, starting at the end of the eighth season, thus at the end of the observation period, and ending at the end of the tenth season, that is, the season which is two seasons after the observation period. 

The true GE values were approximated based on $2 \cdot 10^5$ observations, using the entirety of the available training data for training, that is, the first eight seasons. The number of training observations was also varied ($\ntrain \in \{100, 500, 1000, 3000\}$). As in the simulations presented in previous sections, linear models and random forests were used as learners, and MSE as the error measure.

\begin{figure}
  \centering
\includegraphics[width = \textwidth]{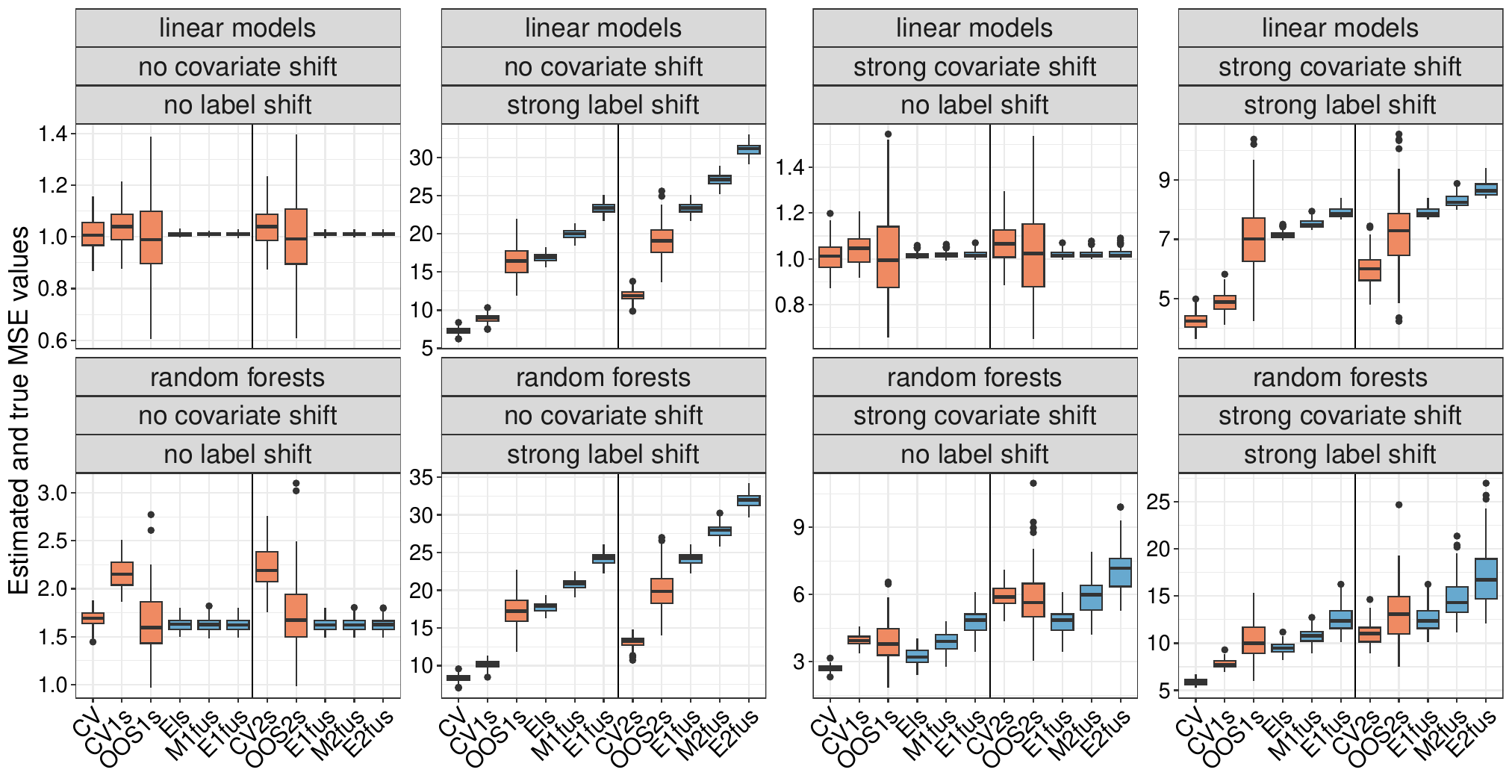}
\caption{Simulation on concept drift: estimated and (approximated) true MSE values. The figure shows the subset of settings with $\ntrain = 500$, focusing only on no or strong feature and label shifts. The orange and blue boxplots indicate the estimated and true MSE values, respectively. {\rcode{CV}} refers to the results obtained for (standard) CV. {\rcode{CV1s}} and {\rcode{CV2s}} represent time-series CV without and with a buffer zone. {\rcode{OOS1s}} and {\rcode{OOS2s}} stand for out-of-sample validation without and with a buffer zone. {\rcode{Els}}, {\rcode{M1fus}}, {\rcode{E1fus}}, {\rcode{M2fus}}, and {\rcode{E2fus}} indicate the true errors at the end of the eighth season, the middle and end of the ninth season, and the middle and end of the tenth season, respectively.}
\label{fig:concdriftsim}
\end{figure}

\paragraph{Results and discussion}
Figure~\ref{fig:concdriftsim} provides a visualization of the estimated and true MSE values for a representative subset of the simulation settings for $\ntrain = 500$. The settings depicted focus on cases of either no feature shift or strong feature shift, as well as cases of either no label shift or strong label shift. It is worth noting that while this figure only covers a subset of all simulation settings, it is nonetheless quite representative, as the omitted settings (i.e., weak and medium-strong label and feature shifts) can be seen as intermediate stages between the no-shift and strong-shift settings depicted here. Additionally, the training set size $\ntrain$ did not notably affect the results. For a complete overview of all the simulation results and corresponding descriptions, please refer to Sections~B.2 and B.3 of the Supplementary Materials.

CV and time-series CV (i.e., prequential validation) generally provided over-optimistic GE estimates in most settings. Out-of-sample validation produced mostly unbiased estimates for the immediate post-observation period. However, the GE of the trained models tended to increase rapidly after the observation period. This suggests that it is important to retrain the models frequently, re-estimating the GE with out-of-sample validation. Using out-of-sample validation with a temporal buffer zone between training and test data, the GE in the slightly more distant future could be estimated realistically in many cases. However, in some settings, we observed over-optimistic error estimates, although this over-optimism was not very strong. Out-of-sample validation was associated with a relatively large variance, especially for small training set sizes, making it difficult to estimate the GE realistically for small data sets. However, the strong biases of CV and time-series CV were much more pronounced, which is why the error estimates from out-of-sample validation were much closer to the GEs despite their greater variance.

An exception where CV did not lead to over-optimism occurred in settings with pure feature drift, but only for linear models. This is because, as discussed in Section~\ref{sec:concdrifttypes}, linear models can extrapolate well to regions of the feature space not contained in the training data---if the feature influences are actually linear, as was the case in the simulation. In contrast, random forests provided over-optimistic error estimates even for pure feature drift, as they are not suited for extrapolating to unseen regions of the feature space. As also indicated in Section~\ref{sec:concdrifttypes}, pure feature shift generally seems to be of little concern in the literature, but our results suggest that this may not be warranted.

In settings with no or weak concept drift, time-series CV produced notably pessimistic error estimates. This was also indicated by \citet{Gama:2013}, who argued that this can happen because the first model evaluations are obtained based on small training set sizes. This is a more severe form of the bias of the holdout estimator discussed in Section~\ref{subsec:resampling}.

It is not very surprising that out-of-sample validation had the lowest bias since, with this procedure, the training data is very similar to the entire data set $\D$ on which the final model is fitted, and the test data, being at the end of the observation period, is very similar to the next observations after the observation period. This is also the case for other forms of concept drift beyond the incremental drift examined in the simulation, such as abrupt drift, which is why out-of-sample validation should work well for these types of drifts as well. In general, it appears difficult to set up realistic concept drift scenarios in simulations because there is no real data analysis-based overview of how concept drift typically manifests in applications \citep{Lu:2018}.

\subsubsection{Conclusions for concept drift}
\label{sec:concdriftconcl}
Changes in the distribution of data received in streams, known as concept drift, can cause the prediction performance of ML models to decline if not properly addressed. Detecting concept drift and implementing appropriate responses is therefore crucial. Numerous methodologies are available for these tasks. However, the existing literature concerning the estimation of the GE under concept drift is relatively sparse.

Prequential validation, also referred to as time-series CV, is the prevalent method for estimating the GE in the context of concept drift. Nonetheless, our simulation study indicates that both prequential validation and standard CV can yield overoptimistic GE estimates. In contrast, out-of-sample validation, which uses only the most recent observations for GE estimation, appears to produce largely unbiased estimates.

Introducing a temporal buffer between the training and test data allows for the estimation of the GE expected slightly further into the future. However, the GE can rise rapidly, underscoring the necessity for frequent retraining of the ML model. A drawback of out-of-sample validation is that it can result in a relatively high variance in the GE estimate when the number of cases is small. Yet, considering the pronounced bias expected when using alternative methods, we still advocate for the estimation of the GE under concept drift using out-of-sample validation.

Lastly, the use of standard CV under conditions of pure feature drift, due to extrapolations, seems to also lead to underestimation of the GE, contrary to commonly held beliefs in the concept drift literature. An important exception, where no bias is expected, includes correctly specified classical regression models that linearly model feature influences, given the true feature influences are linear as well.

\subsection{Hierarchical classification}
\label{subsec:hiercl}

\subsubsection{Types of hierarchical classification problems}
\begin{figure}
\centering
\includegraphics[width = 0.4 \textwidth]{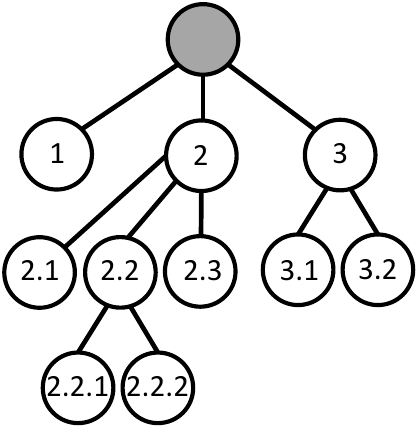}
\caption{Example of a category tree in hierarchical classification problems. Nodes higher up the tree represent more general classes.}
\label{fig:categorytree}
\end{figure}

As we mentioned in the introduction, hierarchical classification problems involve observations that belong not to single classes but rather to collections of nested classes. The relationships among these classes can usually be represented by a category tree \citep{sun.hierarchical.2001}, as illustrated in Figure~\ref{fig:categorytree}.\footnote{There are also hierarchical classification problems where the (sub-)classes form a directed acyclic graph structure, allowing for multiple parent nodes for each node. However, our focus in this paper is on tree-structured classification problems, as displayed in Figure~\ref{fig:categorytree}.} The nodes that mark the most specific class categorizations are called \emph{leaf nodes}. Referring to the ISCO-08 hierarchical classification scheme introduced earlier, the profession \say{fortune-teller} would be a leaf node class arranged hierarchically as follows: 1) Major Group - \say{5 Service and sales workers}; 2) Sub-Major Group - \say{51 Personal service workers}; 3) Minor Group - \say{516 Other personal services workers}; 4) Unit Group - \say{5161 Astrologers, fortune-tellers and related workers}; 5) Occupation - \say{5161.2 Fortune-tellers}.

Hierarchical classification problems can be categorized into single-label and multi-label types. The former assigns each observation to a single (leaf) node, whereas the latter allows an observation to belong to more than one node. Our emphasis will be on single-label classification problems.

Often, many leaf nodes (or, in general, nodes further down in the tree hierarchy) are represented by only a small number of observations. In these situations, it may be more practical to predict these more specific nodes only when the estimated accuracy is sufficiently high. Otherwise, internal node predictions are preferred. This approach is commonly referred to as \emph{optional leaf-node prediction}. However, in many applications, internal node predictions are insufficient, and leaf-node predictions are mandatory. Such situations are known as \emph{mandatory leaf-node prediction} problems \citep{costa.review.2007}. 
We will come back to this topic when we discuss loss functions for hierarchical classification in Section~\ref{subsec:hiereval}. 

\subsubsection{Approaches to hierarchical classification: types and examples}

Hierarchical classification can be approached in five different ways:
\begin{enumerate}
\item Apply multi-class classification to the collection of all leaf nodes without considering the hierarchical structure. Referring to Figure~\ref{fig:categorytree}, a single classifier would be tasked with differentiating between classes 1, 2.1, 2.2.1, 2.2.2, 2.3, 3.1, and 3.2.
\item Independently apply local multi-class classification at each level of the hierarchy. For instance, in Figure~\ref{fig:categorytree}, one classifier would discriminate between classes 1, 2, and 3; another would differentiate between classes 2.1, 2.2, 2.3, 3.1, and 3.2; and a third one would distinguish between classes 2.2.1 and 2.2.2.
\item Apply local multi-class classifiers to distinguish between the child nodes. In the context of Figure~\ref{fig:categorytree}, one classifier would differentiate between classes 1, 2, and 3; a second one between classes 2.1, 2.2, and 2.3; a third between classes 3.1 and 3.2; and a fourth one between classes 2.2.1 and 2.2.2.
\item Apply local binary classifiers at each tree node to differentiate the node's class in a \say{one-vs-rest} style from all other classes at the same level. For example, Figure~\ref{fig:categorytree} would require one classifier to distinguish Class 1 from classes 2 and 3, another to distinguish Class 2 from classes 1 and 3, a third to distinguish Class 3 from classes 1 and 2, a fourth to distinguish Class 2.1 from classes 2.2, 2.3, 3.1, and 3.2, and so forth. An advantage of this approach over approach 3 is that more training observations are available for each classifier.
\item Implement global classifiers, often called \say{big-bang} approaches, which consider the entire hierarchical structure.
\end{enumerate}

With the exception of approaches 1 and 2, predictions are performed in a top-down fashion. This means starting from the root node, the test observation is assigned to the most likely child nodes until a leaf node is reached. An issue with this procedure is the propagation of classification errors from higher levels of the tree to lower ones \citep{costa.review.2007}. Generally, the first two approaches are not favored because they disregard the hierarchical structure and, in the case of the second approach, may lead to inconsistencies.

Any binary or multi-class classification methods can be used for the first four approaches. The \say{big-bang} approaches, on the other hand, are classification methods specifically designed for hierarchical classification problems. We will provide an overview of a few common \say{big-bang} approaches below. Since this part is not critical for understanding the remainder of this subsection, readers, depending on their interest, may skip ahead to Section~\ref{subsec:hiereval}.

\citet{zhou.hierarchical.2011} proposed a hierarchical support vector machine (SVM) for tree-structured classification problems with local classifiers per level. This method encourages classifiers within the same tree branch to be distinct by regularizing the SVM weights. This is beneficial because different features are generally important at different levels of the tree hierarchy. The method also encourages more complex classifiers at lower tree levels since \citet{zhou.hierarchical.2011} argue that classification tends to be more challenging in these areas.

Interestingly, \citet{bennett.refined.2009} suggest an approach that uses more complex classifiers at the tree's upper levels. They argue that the classification problems are more complex at these levels due to more general classes, involving a variety of topics, potentially leading to highly non-linear optimal decision boundaries between classes. They also address the error propagation issue in hierarchical classifiers, where errors made higher up in the tree cascade down. They attempt to mitigate this by assigning the training data to the nodes based on cross-validated class predictions rather than the actual classes.  Here it is assumed that the training data allocated this way will better resemble data encountered during the prediction phase, lessening the impact of error propagation.

Contrary to the approach by \citet{zhou.hierarchical.2011}, the method by \citet{gopal.recursive.2013} encourages the parameters from close categories in the hierarchy to be similar by penalizing the differences between each node's parameters and its parents. This aims to obtain more stable parameter estimates for smaller nodes. \citet{gopal.recursive.2013} present two versions of their approach: one based on SVMs and the other on regularized logistic regression.

\citet{mccallum.improving.1998} proposed a Naïve Bayes-type algorithm for tree-structured classification problems with binary input features (presence vs.\ absence of words in text documents). Like \citet{gopal.recursive.2013}, they reduce the variance of parameter estimates at lower levels of the tree hierarchy by shrinking them towards the values of parameter estimates at higher levels. \citet{charuvaka.hiercost.2015} argue that shrinking, as performed by \citet{mccallum.improving.1998} and \citet{gopal.recursive.2013}, has the beneficial effect of lessening the importance of misattributing observations to classes near their true classes in the hierarchy. They achieve the same effect by incorporating hierarchical information into the loss during learning. Specifically, they consider one versus all classifiers at the leaf nodes and weight the observations' contributions to the losses by the dissimilarities between their true classes and the classes associated with the respective classifiers. Interestingly, their approach also allows for multi-label classification.

For large-scale hierarchical classification problems involving hundreds of thousands of categories, training a single classifier becomes computationally infeasible. \citet{xue.deep.2008} proposed an approach for such situations. It is based on training different classifiers for each new observation, using only categories likely associated with the respective observation according to a screening procedure.

In recent years, various hierarchical classification approaches based on (deep) neural networks have been published. These approaches have been predominantly used in text and image categorization \citep{kowsari.hdltex.2017, peng.largescale.2018, wehrmann.hierarchical.2018, gargiulo.deep.2019, huang.hierarchical.2019, mao.hierarchical.2019, aly.hierarchical.2019}.

\subsubsection{Evaluation metrics for hierarchical classification problems}
\label{subsec:hiereval}

This section offers a comprehensive description of evaluation metrics typically used in hierarchical classification problems, which differs from standard measures outlined in Section~\ref{sec:general}. In what follows, we will represent the predicted class as $\widehat{y}^{(i)}$ and, as in Section~\ref{sec:general}, the corresponding true class will be denoted by $\yi$. The latter is always a leaf node class, while $\widehat{y}^{(i)}$ is the deepest class predicted for observation $i$ in the hierarchy of the tree (which, in case of mandatory leaf-node prediction is again always a leaf node class). It is crucial to remember that $\widehat{y}^{(i)}$ does not (necessarily) equate to the prediction $\fh(\xi)$ as defined in Section~\ref{sec:general}. This discrepancy occurs because $\fh$ generates score values (such as predicted class probabilities) that can be converted into $\widehat{y}^{(i)}$ using a specific decision rule (such as selecting the class with the highest predicted probability).

In hierarchical classification, the degree of misclassification severity is influenced by the dissimilarity between the predicted class and the actual class. For instance, misclassifying the actual Class~1.1.1 as 2.1.1 is usually far more consequential than misclassifying it as 1.1.2. Traditional multi-class evaluation metrics such as precision, recall, and the $F_1$ score, do not take this into account because they assign the same loss to all misclassifications, independent of the distance of the predicted class from the true one. Yet, these metrics remain popular for hierarchical classification problems, hence we include them in the following.

Precision {\em Pr} and recall {\em Re} can be computed in two distinct ways for multi-class outcomes; these are the micro-average and the macro-average \citep{Sokolova:2009}:
\begin{enumerate}
    \item Micro-average:
    \begin{equation}
    Pr^\mu = \frac{\sumin |\yi \cap \widehat{y}^{(i)}|}{\sumin |\widehat{y}^{(i)}|}, \quad    Re^\mu = \frac{\sumin |\yi \cap \widehat{y}^{(i)}|}{\sumin |\yi|}
    \end{equation}
    \item Macro-average:
    \begin{equation}
\label{eq:ges}
\begin{aligned}
& Pr^M = \frac{1}{|\Yspace|} \sum\limits_{l \in \Yspace} Pr_l, \quad Re^M = \frac{1}{|\Yspace|} \sum\limits_{l \in \Yspace} Re_l, \quad \text{where} \\
&\quad Pr_l = \frac{\sumin \I(\widehat{y}^{(i)} = l) \;\; |\yi \cap \widehat{y}^{(i)}|}{\sumin \I(\widehat{y}^{(i)} = l) \;\; |\widehat{y}^{(i)}|} \quad \text{and} \\
&\quad Re_l = \frac{\sumin \I(\yi = l) \;\; |\yi \cap \widehat{y}^{(i)}|}{\sumin \I(\yi = l) \;\; |\yi|}.
\end{aligned}\nonumber
\end{equation}
\end{enumerate}
In this context and in the following, $|\cdot|$ again represents the cardinality. Note that in the above formulas, since $\yi$ and $\widehat{y}^{(i)}$ are scalar, we could have written $\I(\widehat{y}^{(i)} = \yi)$ instead of $|\yi \cap \widehat{y}^{(i)}|$ and $1$ instead of $|\yi|$ and $|\widehat{y}^{(i)}|$. However, since we will need the set-based notation later in the definitions of the hierarchical versions of the above measures, we already introduce it here for consistency.

The micro-average assigns equal weights to all observations, while the macro-average assigns equal weights to all $\vert \Yspace\vert$ classes. Precision and recall should only be calculated in the case of mandatory leaf-node prediction problems. Given that $|\widehat{y}^{(i)}| = |\yi| = 1$ and therefore $\sumin |\widehat{y}^{(i)}| = \sumin |\yi| = n$, it is straightforward to observe that both $Pr^\mu$ and $Re^\mu$ are equivalent to the accuracy. Moreover, $Pr_l$ represents the proportion of correct predictions within the predictions of Class~$l$, and $Pr^M$ denotes the average of the $Pr_l$ values across all classes in $\Yspace$. Similarly, $Re^l$ is the proportion of correct predictions within observations from Class~$l$, and $Re^M$ represents the average of the $Re^l$ values across all classes in $\Yspace$. Please note that there is a slight abuse of notation in the formula for $Pr^M$; to avoid dividing by zero, only the leaf node classes predicted at least once should be included in the calculation instead of $\Yspace$.

Given that precision and recall assess complementary aspects of classification performance, the $F_1$ score that combines both is frequently considered. It is defined as the harmonic mean of precision and recall, and can be calculated using either the micro-average or macro-average versions of precision and recall, as defined above \citep{sun.hierarchical.2001}.

\citet{kiritchenko.functional.2005} introduce versions of the aforementioned measures that accommodate the hierarchical nature of the classification problem. Specifically, these measures respect the fact that each class also pertains to all ancestor classes; for instance, Class~2.1.3 belongs to Class~2.1 and Class~2. The versions extend the predicted and true classes $\widehat{y}^{(i)}$ and $\yi$ by the sets of all ancestors of these classes (see Figure~\ref{fig:categorytree}), thus generating two expanded sets $\widehat{\boldsymbol{y}}_{AUG}^{(i)}$ and $\boldsymbol{y}_{AUG}^{(i)}$. Subsequently, in the above formulas, they replace $|\yi \cap \widehat{y}^{(i)}|$, $|\widehat{y}^{(i)}|$, and $|\yi|$ by $|\boldsymbol{y}_{AUG}^{(i)} \cap \widehat{\boldsymbol{y}}_{AUG}^{(i)}|$, $|\widehat{\boldsymbol{y}}_{AUG}^{(i)}|$, and $|\boldsymbol{y}_{AUG}^{(i)}|$, respectively. The resulting measures are referred to as hierarchical precision, hierarchical recall, and hierarchical $F_1$ score. These measures can also be applied in scenarios involving optional leaf-node prediction problems. In such cases, for the calculation of  the macro-averaged hierarchical precision, all classes that have been predicted at least once are considered, whether they are internal or leaf node classes, as opposed to considering $\Yspace$.

\citet{sun.hierarchical.2001} concentrate exclusively on text classification and propose a different approach for extending precision and recall for hierarchical classification. Like \citet{kiritchenko.functional.2005}, they do not reward exact predictions exclusively, but also predictions with high similarity to the true classes. The higher this similarity, the larger the assigned rewards, which then are added to the numerator and denominator in the precision and recall formulas (either micro-averaged or macro-averaged version). The difference to \citet{kiritchenko.functional.2005} is that \citet{sun.hierarchical.2001} assess class similarity based solely on data associated with the corresponding classes, not on the hierarchical relationship between the classes. More specifically, class similarity is  measured based on the similarities of the words in the documents affiliated with the different classes. The fact that this approach does not exploit the hierarchy tree structure has been cited as a drawback in \citet{freitas.tutorial.01}. Yet, \citet{sun.hierarchical.2001} also examine a version of their approach that uses the distances between the classes in the tree.

\citet{cai.exploiting.2007} employ the symmetric difference loss between the predicted and true classes, considering again the classes augmented with their ancestor sets:
\begin{equation}
\frac{1}{n} \sumin \left|\left\{\widehat{\boldsymbol{y}}_{AUG}^{(i)} \setminus \boldsymbol{y}_{AUG}^{(i)}\right\} \cup \left\{\boldsymbol{y}_{AUG}^{(i)} \setminus \widehat{\boldsymbol{y}}_{AUG}^{(i)}\right\}\right|.
\end{equation}
This corresponds to the union of the false positives and false negatives when considering the classes augmented with their ancestor sets \citep{kosmopoulos.evaluation.2015}.

\citet{wang.building.1999} develop their loss measure based on the hierarchical classification problem's geometry (see Figure~\ref{fig:categorytree}). They calculate the number of edges on the shortest path between the predicted class $\widehat{y}^{(i)}$ and the true class $\yi$ for each observation. Notably, for tree-structured hierarchical classification problems, this loss measure corresponds to the symmetric difference loss by \citet{cai.exploiting.2007}.

One shortcoming of the aforementioned loss measure in optional leaf-prediction scenarios is that it fails to acknowledge that misclassifications at the hierarchy's lower levels are less harmful than those at higher levels. \citet{blockeel.hierarchical.2002} also measure the shortest path between the predicted and the actual class but assign level-specific weights to the edges in the hierarchy, which diminish, potentially exponentially, at lower hierarchy levels.

As mentioned in the previous subsection, with top-down classification methods, errors made at the top of the tree propagate downwards. Consequently, once an observation is incorrectly classified for the first time in the top-down process, any subsequent classifications for that observation are likely to be incorrect. This led \citet{cesa-bianchi.incremental.2004} to introduce the H-loss, a loss measure that only penalizes the initial errors in top-down classification. Let $y_{AUG, 1}^{(i)}, \dots, y_{AUG, L}^{(i)}$ and $\hat{y}_{AUG, 1}^{(i)}, \dots, \hat{y}_{AUG, L^*}^{(i)}$ represent the elements of $\boldsymbol{y}_{AUG}^{(i)}$ and $\widehat{\boldsymbol{y}}_{AUG}^{(i)}$, respectively, where the first and last elements correspond to the most general and the most specific classes, correspondingly, in the tree hierarchy. The H-loss is then defined as:
\begin{equation}
\frac{1}{n} \sumin \sum_{l=1}^{L^*} c_l \; \I\Big(\hat{y}_{AUG, l}^{(i)} \neq y_{AUG, l}^{(i)} \;\; \land \;\; \left[\hat{y}_{AUG, l^*}^{(i)} = y_{AUG, l^*}^{(i)} \;\; \forall \;\; l^* < l\right]\Big),
\end{equation}
where $c_1 > \dots > c_L > 0$ are fixed cost coefficients. These coefficients decrease at lower hierarchy levels to reflect the generally lesser impact of errors in this region. \citet{blockeel.hierarchical.2002} did not only present the measure described in the last paragraph, but also proposed a second measure, assigning weights to the nodes rather than the edges and taking the weight of the deepest common ancestor of the predicted and the true class as the loss measure. This measure can be seen as a specific case of the measure proposed by \citet{cesa-bianchi.incremental.2004}.

We can calculate the loss $\Lxy$ associated with a specific loss measure for each potential pair of predicted and true classes in $\Yspace$. \citet{freitas.tutorial.01} suggest presenting these misclassification losses in a matrix with true and predicted classes in the columns and rows, respectively, that is, to use a misclassification cost matrix. This approach can facilitate comparisons between different loss measures, but it may become challenging to interpret as the number of classes (and therefore the dimension of the matrix) increases.

\citet{wu.hierarchical.2019} propose a method for evaluating probability predictions. Here, it is assumed that the classifier assigns a probability to each leaf node class such that the probabilities sum to one. By aggregating these probabilities, we can readily assign probabilities to each internal class node. \citet{wu.hierarchical.2019} measure classification rewards instead of losses, coining their measure as \say{win}. This win is computed by dropping the observation down the hierarchy tree, adding $0.5^m$ times the probability assigned to the corresponding (internal) node to the win at the $m$th node (the first node being the root node, starting with zero win). This process stops before reaching the first incorrectly classified (internal) node. If all encountered node classes, including the leaf node class, are classified correctly, the win contribution of the leaf node is added twice. This amplifies the reward for correctly predicting the leaf node class over a prediction that is correct up until the node before the leaf node. With this measure, more confident probability predictions receive higher rewards. Notably, for non-probabilistic predictions (i.e., the predicted leaf node's probability is one), the procedure by \citet{wu.hierarchical.2019} is a special case of \citet{blockeel.hierarchical.2002}, and hence also of \citet{cesa-bianchi.incremental.2004}.

For a comprehensive overview on many of the measures described above, refer to \citet{costa.review.2007} and \citet{kosmopoulos.evaluation.2015}. For single-label and tree-structured problems, pair-based strategies like those by \citet{cesa-bianchi.incremental.2004}, \citet{blockeel.hierarchical.2002}, and \citet{wang.building.1999} are suitable. Hierarchical precision, hierarchical recall, and hierarchical $F_1$ score have the drawback of overpenalizing errors in nodes with many ancestors \citep{kosmopoulos.evaluation.2015}. Besides reviewing existing measures, \citet{kosmopoulos.evaluation.2015} introduce two new measures, one of which avoids this overpenalization issue. These measures can also handle multi-label classification problems and directed acyclic graphs. However, due to their complex nature, they are not discussed further here.

\subsubsection{Resampling for hierarchical classification problems}
To the best of our knowledge, existing literature does not provide specific guidelines on selecting resampling methods for performance estimation in hierarchical classification problems. Nevertheless, we propose that stratified CV \citep{Breiman:1984} might be a more favorable choice compared to standard CV when dealing with hierarchical outcomes, as we will detail subsequently. An empirical study conducted by \citet{kohavi.study.1995}, which examined various resampling methods for error estimation on six data sets with binary and multi-class outcomes, concluded that stratified CV offers superior performance over standard CV in terms of bias and variance. Given that hierarchical classification typically deals with a large number of classes, we theorize that the advantages of stratifying the folds in CV with respect to the class distributions, as observed by \citet{kohavi.study.1995}, could be particularly notable in the context of hierarchical classification. However, it is not immediately obvious how the folds should be stratified in relation to the classes in hierarchical classification problems. In traditional multi-class classification, the folds can be stratified such that each fold maintains the same empirical class distribution. In contrast, hierarchical classification presents more complex class structures (see Figure~\ref{fig:categorytree}). Yet, it becomes evident that stratifying the folds based on the collective leaf node classes ensures the folds are stratified according to the class distributions at each level of the hierarchy in the tree. Occasionally, leaf node classes are represented by only a handful of observations. Specifically, when a class has fewer observations than the folds $K$ in CV, it becomes impossible to distribute observations from that class to each fold. In such instances, each of these observations can be randomly assigned to a different one of the $K$ folds.

In the next subsection, we present a small simulation study that reinforces our conjecture that stratified CV may be preferable for hierarchical classification problems, albeit the differences in the results between the two procedures were relatively small.

\subsubsection{Simulation study comparing stratified with non-stratified CV for hierarchical classification problems}
\label{subsec:hiersimul}

\paragraph{Objective, study design and simulation model}
We evaluated standard and stratified 5-fold CV, both repeated ten times, in terms of their ability to estimate the following performance metrics: accuracy, micro-averaged / macro-averaged hierarchical precision, micro-averaged / macro-averaged hierarchical recall, micro-averaged / macro-averaged hierarchical $F_1$ score, weighted shortest path loss measure, and H-loss.

We generated a single category tree with 50 leaf node classes and 39 internal class nodes, with each internal node having either two or three child nodes. This tree was used in all simulation iterations. We considered five features with node-specific effects that weakened at lower tree levels, making child node data more similar in the lower tree layers---a common characteristic of hierarchical classification problems. The feature values were sampled i.i.d.\ from $\normal(0,1)$. For simulating the outcome, the observations were dropped down the category tree. At each node, observations were allocated to its child nodes using a multinomial regression model. We considered different sample sizes ($\ntrain \in \{200, 500, 1000, 3000\}$) and approximated the true evaluation metric values on test data sets of size $\ntest = 2 \cdot 10^5$, where the entire available data set $\D$ was used for training. As learners, we used top-down hierarchical classification with random forests as local multi-class classifiers  to distinguish between the child nodes, implemented in our R package \say{hierclass} available on GitHub (version 0.1.0, link: \url{https://github.com/RomanHornung/hierclass}). A detailed description of the simulation design and the results is provided in Section~C of the Supplementary Materials.

\paragraph{Results}
Figure~\ref{suppfig:standardizeddifference} shows the (standardized) differences between the evaluation metrics estimates and their true values. In summary, for very small sample sizes, stratified CV slightly overestimated the true performance, whereas standard CV slightly underestimated it. For larger sample sizes, however, the negative bias of standard CV was reduced but persisted, whereas stratified CV exhibited no notable bias. The variance of the estimates did not notably differ between the two procedures. However, the results for the macro-averaged performance metrics differed, exhibiting a noticeable optimistic bias for both CV methods, which was more pronounced for small sample sizes. Given that these metrics assign equal weight to all classes, this bias could be attributable to the fact that the training data sets $\D$ tended to lack many small classes, especially for small training set sizes.

\begin{figure}
  \centering
\includegraphics[width = \textwidth]{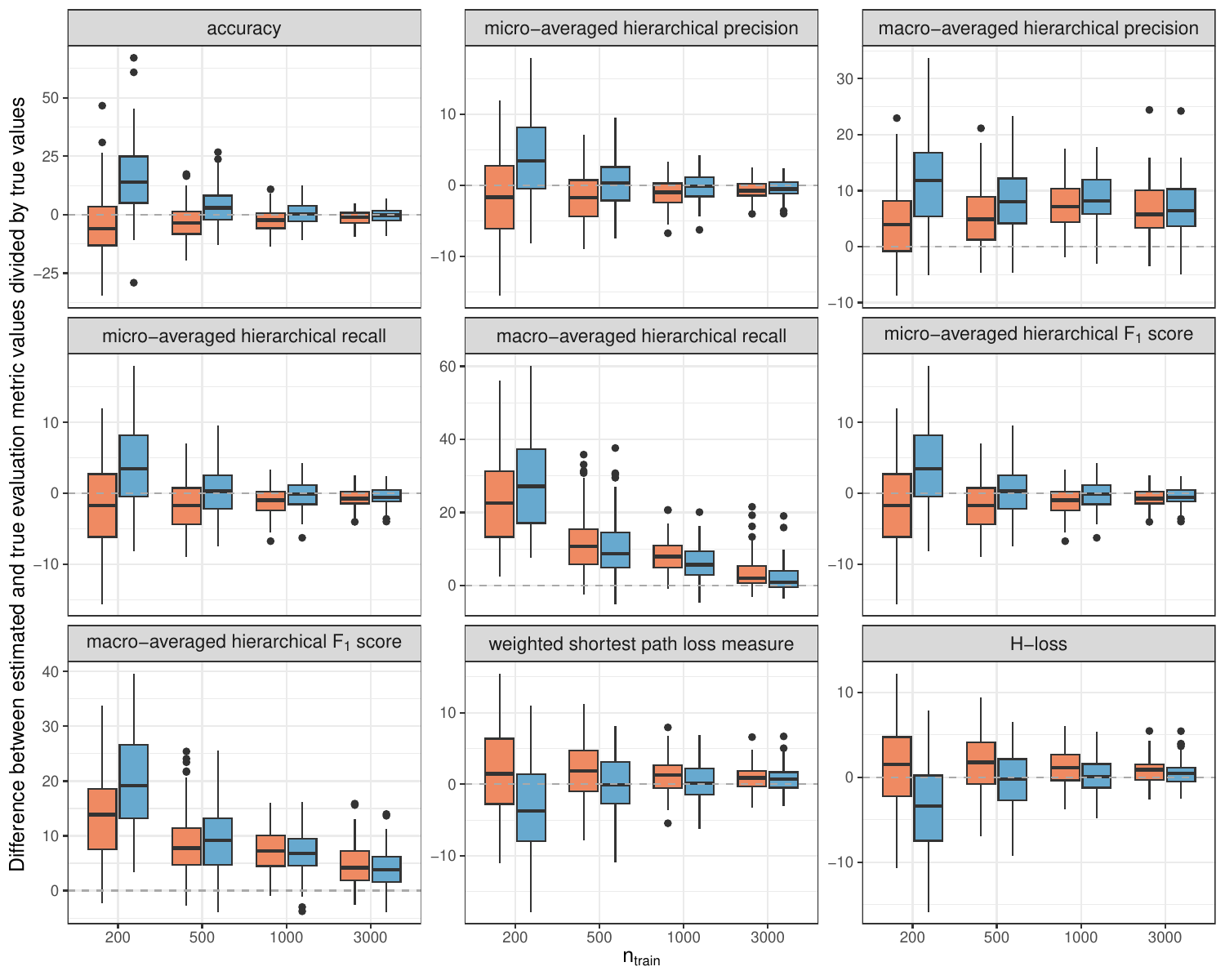}
\caption{Simulation on hierarchical classification: differences between estimated and true evaluation metric values divided by true values. The orange and the blue boxplots indicate the results obtained when using standard and stratified CV, respectively.}
\label{suppfig:standardizeddifference}
\end{figure}

\subsubsection{Conclusions for hierarchical classification}
\label{sec:hierclconcl} 
Hierarchical classification problems differ from conventional classification problems in that each observation belongs to a set of nested classes rather than a single class. The relationships between all classes can be visualized through a category tree. ML methods for such problems can be divided into those that repeatedly use identical learners to make decisions at each point in the category tree, and so-called \say{big-bang} approaches that tackle the structure of the hierarchical classification problem as a whole.

Performance measures for hierarchical classification problems are designed to consider that the severity of misclassification is determined by the extent to which the predicted class deviates from the true class. A variety of such measures exist, each with its unique advantages and limitations. Additionally, these measures often vary in their compatibility with different types of hierarchical classification problems.

To our knowledge, existing literature has not investigated which resampling techniques are best suited for hierarchical classification problems. We conducted a simulation study to test our hypothesis that stratified CV results in lower bias and variance compared to standard CV. While both methods showed some bias for small data sets, stratified CV showed negligible bias with increasing sample sizes. Hierarchical classification problems typically require relatively large data sets. This characteristic makes stratified CV a more reliable method for performance estimation in such contexts. Therefore, we advocate the use of stratified CV for performance estimation of hierarchical classification models. Our simulation study did not reveal notable differences in variance between stratified and standard CV.

\section{Discussion and overall conclusions}
\label{sec:discconcl}
This paper reviewed both pre-existing literature on the subject of GE estimation in non-standard settings, with a particular focus on different types of non-i.i.d.\ situations, and additionally provided new insights derived from simulation studies. By integrating the established knowledge from the literature with new findings, we aim to provide a cohesive understanding, serving as a foundation for comprehensive and informed guidance.

A universal principle applicable in GE estimation in non-standard settings is that the test data used in the resampling procedure should reflect the new observations to which the model will be applied, while the training data $\Dtrain$ should correspond closely to the entire data set $\D$ on which the final model is obtained. This is evidenced for all five settings reviewed in this paper. For instance, with clustered data intended for predictions on new clusters, it should be ensured that, within CV, the test data contain different clusters than the training data, which is achieved by assigning observations to folds on a cluster-by-cluster basis. In the context of spatial CV, the imposed separation degree between $\Dtrain$ and $\Dtest$ should reflect the distance from the available observations to the new data for which predictions are being made. To adjust the GE estimator for unequal sampling probabilities, the test data observations are reweighted to reflect the population distribution for which predictions are intended. In the case of streaming data, which might exhibit concept drift, the ML model is applied to observations post the observation period, and thus, data towards the end of the observation period should be used as $\Dtest$. In hierarchical classification problems with many small classes represented by few observations, our recommended stratified CV ensures $\Dtrain$ mirrors the class distribution of $\D$, eliminating the risk, unlike standard CV, of the proportions of small classes in $\Dtest$ and $\Dtrain$ diverging strongly from those in $\D$ due to randomness.

The general principle that the structure of the training and test data in resampling procedures for non-standard settings should be shaped by the prediction scenario in question can guide the development of suitable resampling methods for other non-standard scenarios not addressed in this paper, such as time series or spatio-temporal data. However, this principle is merely suggestive of potential resampling methods. Validation of new resampling methods and information on their optimal configurations necessitates empirical and, if feasible, theoretical results.

We have only explored the choice of resampling method for model selection and tuning parameter optimization within the context of the non-standard settings \say{spatial data} and, to a considerably smaller extent, \say{clustered data}. This focus is due to the lack of literature, as per our knowledge, that examines these issues for other non-standard settings. However, as already highlighted in Section~\ref{subsec:scvmodelsel}, these topics would also be relevant for the other non-standard settings we have treated. The results of our simulation studies indicated that the use of standard resampling methods also leads to biased GE estimates in the other non-standard settings considered. It is possible that this bias varies among different model types or within the same model type depending on the tuning parameter values. If this is the case, it would be crucial to choose an appropriate resampling method in model selection or tuning parameter optimization for these non-standard settings. Further research is needed to clarify this point. However, to err on the side of caution and to maintain consistency, we recommend the use of the resampling methods suggested in this paper for the respective non-standard settings in both model selection and tuning parameter optimization, mirroring the advice given by \citet{Schratz:2021}, \citet{Gholamiangonabadi:2020}, and \citet{Saeb:2017} in the context of spatial and clustered data.

Our simulation studies were meticulously designed to hopefully enable unbiased investigation of the issues they explore. Specifically, we varied the values of critical parameters thought to influence the results and carefully defined the parameter values used. Nevertheless, there are opportunities for further extension of the simulation studies. For instance,  additional parameter values or varying other parameters like the number of features or $k$ in $k$-fold CV could be considered. Non-linear influences or categorical features could also be incorporated. Other ML learners such as neural networks, or in the case of hierarchical classification, other hierarchical learners could have been taken into account, although no such methods seem to be currently implemented in R beyond the top-down approach available in our R package \say{hierclass}.

While the simulations could be expanded as discussed, the obtained results were consistent and generally aligned with our anticipations. We aimed for simulation settings that were only as extensive as needed to avoid confusion in result interpretation. Additionally, as described in the introduction, we designed the data-generating processes to be as similar as possible across the different non-standard settings to increase the generalizability of our findings. To refrain from drawing non-generalizable conclusions that would not stand under different simulation designs, we avoided interpreting result details and made only general conclusions.

In summary, both the literature review and our simulation studies indicate that in non-standard settings, using standard resampling methods to estimate the GE error is typically inadequate, as these are usually associated with biased error estimates for these settings. The insights provided here can assist researchers faced with non-standard settings to realistically estimate the performance of ML models.

\section*{Acknowledgements}
The authors would like to thank Savanna Ratky for valuable language corrections. With the exception of L. Schneider, all authors were supported by the Federal Statistical Office of Germany within the cooperation \say{Machine Learning in Official Statistics}. M. Nalenz received financial and general support from the LMU Mentoring Program. L. Schneider was partially supported by the Bavarian Ministry of Economic Affairs, Regional Development and Energy through the Center for Analytics - Data - Applications (ADACenter) within the framework of BAYERN DIGITAL II (20-3410-2-9-8).

\section*{Supplementary Materials and R code}
The Supplementary Materials can be found at the following link: \url{http://www.ibe.med.uni-muenchen.de/organisation/mitarbeiter/070_drittmittel/hornung/pperfestcomplex_suppfiles/suppmat_pperfestcomplex.pdf}. The R code used to produce the results shown in the main paper and in the Supplementary Materials is available on GitHub (\url{https://github.com/RomanHornung/PPerfEstComplex}).

\bibliographystyle{abbrvnat}
\bibliography{bibliography}

\begin{thebibliography}{106}
\providecommand{\natexlab}[1]{#1}
\providecommand{\url}[1]{\texttt{#1}}
\expandafter\ifx\csname urlstyle\endcsname\relax
  \providecommand{\doi}[1]{doi: #1}\else
  \providecommand{\doi}{doi: \begingroup \urlstyle{rm}\Url}\fi

\bibitem[Aly et~al.(2019)Aly, Remus, and Biemann]{aly.hierarchical.2019}
R.~Aly, S.~Remus, and C.~Biemann.
\newblock Hierarchical multi-label classification of text with capsule
  networks.
\newblock In \emph{Proceedings of the 57th Annual Meeting of the Association
  for Computational Linguistics: Student Research Workshop}, pages 323--330,
  2019.

\bibitem[Austern and Zhou(2020)]{Austern:2020}
M.~Austern and W.~Zhou.
\newblock Asymptotics of cross-validation.
\newblock {arXiv:2001.11111}, 2020.

\bibitem[Bates et~al.(2023)Bates, Hastie, and Tibshirani]{Bates:2023}
S.~Bates, T.~Hastie, and R.~Tibshirani.
\newblock Cross-validation: What does it estimate and how well does it do it?
\newblock \emph{J. Am. Stat. Assoc.}, pages 1--12, 2023.
\newblock \doi{10.1080/01621459.2023.2197686}.

\bibitem[Bayle et~al.(2020)Bayle, Bayle, Janson, and Mackey]{Bayle:2020}
P.~Bayle, A.~Bayle, L.~Janson, and L.~Mackey.
\newblock Cross-validation confidence intervals for test error.
\newblock \emph{Adv. Neur. In.}, 33:\penalty0 16339--16350, 2020.
\newblock \doi{10.1002/sim.9873}.

\bibitem[Bengio and Grandvalet(2004)]{bengio.no.2004}
Y.~Bengio and Y.~Grandvalet.
\newblock No unbiased estimator of the variance of k-fold cross-validation.
\newblock \emph{J. Mach. Learn. Res.}, 5:\penalty0 1089--1105, 2004.

\bibitem[Bennett and Nguyen(2009)]{bennett.refined.2009}
P.~N. Bennett and N.~Nguyen.
\newblock Refined experts: Improving classification in large taxonomies.
\newblock In \emph{Proceedings of the 32nd International ACM SIGIR Conference
  on Research and Development in Information Retrieval}, pages 11--18, 2009.

\bibitem[Bergmeir et~al.(2018)Bergmeir, Hyndman, and Koo]{Bergmeir:2018}
C.~Bergmeir, R.~J. Hyndman, and B.~Koo.
\newblock A note on the validity of cross-validation for evaluating
  autoregressive time series prediction.
\newblock \emph{Comput. Stat. Data An.}, 120:\penalty0 70--83, 2018.
\newblock \doi{10.1016/j.csda.2017.11.003}.

\bibitem[Bernau et~al.(2014)Bernau, Riester, Boulesteix, Parmigiani,
  Huttenhower, Waldron, and Trippa]{Bernau:2014}
C.~Bernau, M.~Riester, A.-L. Boulesteix, G.~Parmigiani, C.~Huttenhower,
  L.~Waldron, and L.~Trippa.
\newblock Cross-study validation for the assessment of prediction algorithms.
\newblock \emph{Bioinformatics}, 30\penalty0 (12):\penalty0 i105--i112, 2014.
\newblock \doi{10.1093/bioinformatics/btu279}.

\bibitem[Bifet and Gavalda(2007)]{Bifet:2007}
A.~Bifet and R.~Gavalda.
\newblock Learning from time-changing data with adaptive windowing.
\newblock In \emph{Proceedings of the 2007 SIAM International Conference on
  Data Mining}, pages 443--448, 2007.

\bibitem[Bischl et~al.(2012)Bischl, Mersmann, Trautmann, and
  Weihs]{bischl.resampling.2012}
B.~Bischl, O.~Mersmann, H.~Trautmann, and C.~Weihs.
\newblock Resampling methods for meta-model validation with recommendations for
  evolutionary computation.
\newblock \emph{Evol. Comput.}, 20\penalty0 (2):\penalty0 249--275, 2012.
\newblock \doi{10.1162/EVCO_a_00069}.

\bibitem[Bischl et~al.(2023)Bischl, Binder, Lang, Pielok, Richter, Coors,
  Thomas, Ullmann, Becker, et~al.]{Bischl:2023}
B.~Bischl, M.~Binder, M.~Lang, T.~Pielok, J.~Richter, S.~Coors, J.~Thomas,
  T.~Ullmann, M.~Becker, et~al.
\newblock Hyperparameter optimization: foundations, algorithms, best practices,
  and open challenges.
\newblock \emph{WIREs Data Min. Knowl.}, 13\penalty0 (2):\penalty0 e1484, 2023.
\newblock \doi{10.1002/widm.1484}.

\bibitem[Bischl et~al.(2024)Bischl, Sonabend, Kotthoff, and Lang]{Bischl:2024}
B.~Bischl, R.~Sonabend, L.~Kotthoff, and M.~Lang, editors.
\newblock \emph{Applied Machine Learning Using mlr3 in R}, chapter~3.
\newblock CRC Press, 2024.
\newblock ISBN 9781032507545.

\bibitem[Blockeel et~al.(2002)Blockeel, Bruynooghe, D{\v z}eroski, Ramon, and
  Struyf]{blockeel.hierarchical.2002}
H.~Blockeel, M.~Bruynooghe, S.~D{\v z}eroski, J.~Ramon, and J.~Struyf.
\newblock Hierarchical multi-classification.
\newblock In \emph{Workshop Notes of the KDD'02 Workshop on Multi-Relational
  Data Mining}, pages 21--35, 2002.

\bibitem[Boulesteix et~al.(2008)Boulesteix, Strobl, Augustin, and
  Daumer]{Boulesteix:2008}
A.~L. Boulesteix, C.~Strobl, T.~Augustin, and M.~Daumer.
\newblock Evaluating microarray-based classifiers: An overview.
\newblock \emph{Cancer Inform.}, 6:\penalty0 77--97, 2008.
\newblock \doi{10.4137/cin.s408}.

\bibitem[Braga-Neto and Dougherty(2004)]{BragaNeto:2004}
U.~M. Braga-Neto and E.~R. Dougherty.
\newblock Is cross-validation valid for small-sample microarray classification?
\newblock \emph{Bioinformatics}, 20\penalty0 (3):\penalty0 374--380, 2004.
\newblock \doi{10.1093/bioinformatics/btg419}.

\bibitem[Breidt and Opsomer(2017)]{breidt.modelassisted.2017a}
F.~J. Breidt and J.~D. Opsomer.
\newblock Model-assisted survey estimation with modern prediction techniques.
\newblock \emph{Stat. Sci.}, 32\penalty0 (2):\penalty0 190--205, 2017.
\newblock \doi{10.1214/16-STS589}.

\bibitem[Breiman et~al.(1984)Breiman, Friedman, Stone, and
  Olshen]{Breiman:1984}
L.~Breiman, J.~Friedman, C.~J. Stone, and R.~A. Olshen.
\newblock \emph{Classification and Regression Trees}.
\newblock CRC Press, New York, 1984.
\newblock \doi{10.1201/9781315139470}.

\bibitem[Brenning(2005)]{Brenning:2005}
A.~Brenning.
\newblock Spatial prediction models for landslide hazards: Review, comparison
  and evaluation.
\newblock \emph{Nat. Hazard Earth Sys.}, 5\penalty0 (6):\penalty0 853--862,
  2005.
\newblock \doi{10.5194/nhess-5-853-2005}.

\bibitem[Brenning(2012)]{brenning.spatial.2012}
A.~Brenning.
\newblock Spatial cross-validation and bootstrap for the assessment of
  prediction rules in remote sensing: The r package sperrorest.
\newblock In \emph{2012 IEEE International Geoscience and Remote Sensing
  Symposium}, pages 5372--5375, 2012.
\newblock \doi{10.1109/IGARSS.2012.6352393}.

\bibitem[Brenning and Lausen(2008)]{brenning2008estimating}
A.~Brenning and B.~Lausen.
\newblock Estimating error rates in the classification of paired organs.
\newblock \emph{Stat. Med.}, 27\penalty0 (22):\penalty0 4515--4531, 2008.
\newblock \doi{10.1002/sim.3310}.

\bibitem[Cai and Hofmann(2007)]{cai.exploiting.2007}
L.~Cai and T.~Hofmann.
\newblock Exploiting known taxonomies in learning overlapping concepts.
\newblock In \emph{Proceedings of the 20th International Joint Conference on
  Artifical Intelligence}, pages 714--719, 2007.

\bibitem[{Cesa-Bianchi} et~al.(2004){Cesa-Bianchi}, Gentile, and
  Zaniboni]{cesa-bianchi.incremental.2004}
N.~{Cesa-Bianchi}, C.~Gentile, and L.~Zaniboni.
\newblock Incremental algorithms for hierarchical classification.
\newblock In L.~K. Saul, Y.~Weiss, and L.~Bottou, editors, \emph{Advances in
  Neural Information Processing Systems}, volume~17, pages 233--240. MIT Press,
  Cambridge, MA, 2004.

\bibitem[Charuvaka and Rangwala(2015)]{charuvaka.hiercost.2015}
A.~Charuvaka and H.~Rangwala.
\newblock Hiercost: Improving large scale hierarchical classification with cost
  sensitive learning.
\newblock In \emph{Proceedings of the 2015 Joint European Conference on Machine
  Learning and Knowledge Discovery in Databases}, pages 675--690, 2015.

\bibitem[Costa et~al.(2007)Costa, Lorena, Carvalho, and
  Freitas]{costa.review.2007}
E.~P. Costa, A.~C. Lorena, A.~C. P. L.~F. Carvalho, and A.~A. Freitas.
\newblock A review of performance evaluation measures for hierarchical
  classifiers.
\newblock In C.~Drummond, W.~Elazmeh, N.~Japkowicz, and S.~A. Macskassy,
  editors, \emph{Evaluation Methods for Machine Learning II: Papers from the
  AAAI-2007 Workshop}, pages 1--6. AAAI Press, 2007.

\bibitem[Dagdoug et~al.(2021)Dagdoug, Goga, and Haziza]{dagdoug2021model}
M.~Dagdoug, C.~Goga, and D.~Haziza.
\newblock Model-assisted estimation through random forests in finite population
  sampling.
\newblock \emph{J. Am. Stat. Assoc.}, 118:\penalty0 1234--1251, 2021.
\newblock \doi{10.1080/01621459.2021.1987250}.

\bibitem[Dawid(1984)]{Dawid:1984}
A.~P. Dawid.
\newblock Present position and potential developments: Some personal views
  statistical theory the prequential approach.
\newblock \emph{J. R. Stat. Soc. Ser. A.-G.}, 147\penalty0 (2):\penalty0
  278--290, 1984.
\newblock \doi{10.2307/2981683}.

\bibitem[Dobbin and Simon(2011)]{dobbin2011optimally}
K.~K. Dobbin and R.~M. Simon.
\newblock Optimally splitting cases for training and testing high dimensional
  classifiers.
\newblock \emph{BMC Med. Genomics}, 4\penalty0 (1):\penalty0 1--8, 2011.
\newblock \doi{10.1186/1755-8794-4-31}.

\bibitem[Ducange et~al.(2021)Ducange, Marcelloni, and Pecori]{Ducange:2021}
P.~Ducange, F.~Marcelloni, and R.~Pecori.
\newblock Fuzzy {Hoeffding} decision tree for data stream classification.
\newblock \emph{Int. J. Comput. Int. Sys.}, 14\penalty0 (1):\penalty0 946--964,
  2021.
\newblock \doi{10.2991/ijcis.d.210212.001}.

\bibitem[Efron and Tibshirani(1997)]{Efron:1997}
B.~Efron and R.~Tibshirani.
\newblock Improvements on cross-validation: The 632+ bootstrap method.
\newblock \emph{J. Am. Stat. Assoc.}, 92\penalty0 (438):\penalty0 548--560,
  1997.
\newblock \doi{10.2307/2965703}.

\bibitem[Ellenbach et~al.(2021)Ellenbach, Boulesteix, Bischl, Unger, and
  Hornung]{Ellenbach:2021}
N.~Ellenbach, A.-L. Boulesteix, B.~Bischl, K.~Unger, and R.~Hornung.
\newblock Improved outcome prediction across data sources through robust
  parameter tuning.
\newblock \emph{J. Classif.}, 38:\penalty0 212--231, 2021.
\newblock \doi{10.1007/s00357-020-09368-z}.

\bibitem[{European Commission}(2023)]{ec2023impact}
{European Commission}.
\newblock The impact of demographic change -- in a changing environment.
\newblock Commission Staff Working Document SWD(2023) 21 final, 2023.

\bibitem[Freitas and de~Carvalho(2007)]{freitas.tutorial.01}
A.~A. Freitas and A.~C. de~Carvalho.
\newblock A tutorial on hierarchical classification with applications in
  bioinformatics.
\newblock In D.~Taniar, editor, \emph{Research and Trends in Data Mining
  Technologies and Applications}, chapter~7, pages 175--208. Idea Group, 2007.

\bibitem[Frias-Blanco et~al.(2016)Frias-Blanco, del Campo-Avila,
  Ramos-Jim{\'e}nez, Carvalho, Ortiz-D{\'\i}az, and
  Morales-Bueno]{FriasBlanco:2016}
I.~Frias-Blanco, J.~del Campo-Avila, G.~Ramos-Jim{\'e}nez, A.~C. Carvalho,
  A.~Ortiz-D{\'\i}az, and R.~Morales-Bueno.
\newblock Online adaptive decision trees based on concentration inequalities.
\newblock \emph{Knowl.-Based Syst.}, 104:\penalty0 179--194, 2016.
\newblock \doi{10.1016/j.knosys.2016.04.019}.

\bibitem[Gama et~al.(2003)Gama, Rocha, and Medas]{Gama:2003}
J.~Gama, R.~Rocha, and P.~Medas.
\newblock Accurate decision trees for mining high-speed data streams.
\newblock In \emph{Proceedings of the Ninth ACM SIGKDD International Conference
  on Knowledge Discovery and Data Mining}, pages 523--528, 2003.

\bibitem[Gama et~al.(2004)Gama, Medas, Castillo, and Rodrigues]{Gama:2004}
J.~Gama, P.~Medas, G.~Castillo, and P.~Rodrigues.
\newblock Learning with drift detection.
\newblock In \emph{Proceedings of the 17th Brazilian Symposium on Artificial
  Intelligence}, pages 286--295, 2004.

\bibitem[Gama et~al.(2013)Gama, Sebastiao, and Rodrigues]{Gama:2013}
J.~Gama, R.~Sebastiao, and P.~P. Rodrigues.
\newblock On evaluating stream learning algorithms.
\newblock \emph{Mach. Learn.}, 90:\penalty0 317--346, 2013.
\newblock \doi{10.1007/s10994-012-5320-9}.

\bibitem[Gama et~al.(2014)Gama, {\v{Z}}liobait{\.e}, Bifet, Pechenizkiy, and
  Bouchachia]{Gama:2014}
J.~Gama, I.~{\v{Z}}liobait{\.e}, A.~Bifet, M.~Pechenizkiy, and A.~Bouchachia.
\newblock A survey on concept drift adaptation.
\newblock \emph{ACM Comput. Surv.}, 46\penalty0 (4):\penalty0 1--37, 2014.
\newblock \doi{10.1145/2523813}.

\bibitem[Gargiulo et~al.(2019)Gargiulo, Silvestri, Ciampi, and
  De~Pietro]{gargiulo.deep.2019}
F.~Gargiulo, S.~Silvestri, M.~Ciampi, and G.~De~Pietro.
\newblock Deep neural network for hierarchical extreme multi-label text
  classification.
\newblock \emph{Appl. Soft Comput.}, 79:\penalty0 125--138, 2019.
\newblock \doi{10.1016/j.asoc.2019.03.041}.

\bibitem[Gholamiangonabadi et~al.(2020)Gholamiangonabadi, Kiselov, and
  Grolinger]{Gholamiangonabadi:2020}
D.~Gholamiangonabadi, N.~Kiselov, and K.~Grolinger.
\newblock Deep neural networks for human activity recognition with wearable
  sensors: Leave-one-subject-out cross-validation for model selection.
\newblock \emph{IEEE Access}, 8:\penalty0 133982--133994, 2020.
\newblock \doi{10.1109/ACCESS.2020.30107150}.

\bibitem[Gon{\c{c}}alves~Jr.\ et~al.(2014)Gon{\c{c}}alves~Jr.\,
  de~Carvalho~Santos, Barros, and Vieira]{Goncalves:2014}
P.~M. Gon{\c{c}}alves~Jr.\, S.~G. de~Carvalho~Santos, R.~S. Barros, and D.~C.
  Vieira.
\newblock A comparative study on concept drift detectors.
\newblock \emph{Expert. Syst. Appl.}, 41\penalty0 (18):\penalty0 8144--8156,
  2014.
\newblock \doi{10.1016/j.eswa.2014.07.019}.

\bibitem[Gopal and Yang(2013)]{gopal.recursive.2013}
S.~Gopal and Y.~Yang.
\newblock Recursive regularization for large-scale classification with
  hierarchical and graphical dependencies.
\newblock In \emph{Proceedings of the 19th {{ACM SIGKDD}} International
  Conference on Knowledge Discovery and Data Mining}, pages 257--265, 2013.

\bibitem[Hastie et~al.(2009)Hastie, Tibshirani, Friedman, and
  Friedman]{Hastie:2009}
T.~Hastie, R.~Tibshirani, J.~H. Friedman, and J.~H. Friedman.
\newblock \emph{The Elements of Statistical Learning: Data Mining, Inference,
  and Prediction}, volume~2.
\newblock Springer, New York, 2009.
\newblock \doi{10.1007/978-0-387-84858-7}.

\bibitem[Hawkins et~al.(2003)Hawkins, Basak, and Mills]{Hawkins:2003}
D.~M. Hawkins, S.~C. Basak, and D.~Mills.
\newblock Assessing model fit by cross-validation.
\newblock \emph{J. Chem. Inf. Comp. Sci.}, 43\penalty0 (2):\penalty0 579--586,
  2003.
\newblock \doi{10.1021/ci025626i}.

\bibitem[Haziza and Beaumont(2017)]{haziza.construction.2017}
D.~Haziza and J.-F. Beaumont.
\newblock Construction of weights in surveys: A review.
\newblock \emph{Stat. Sci.}, 32\penalty0 (2):\penalty0 206--226, 2017.
\newblock \doi{10.1214/16-STS608}.

\bibitem[Heikkinen et~al.(2012)Heikkinen, Marmion, and Luoto]{Heikkinen:2012}
R.~K. Heikkinen, M.~Marmion, and M.~Luoto.
\newblock Does the interpolation accuracy of species distribution models come
  at the expense of transferability?
\newblock \emph{Ecography}, 35\penalty0 (3):\penalty0 276--288, 2012.
\newblock \doi{10.1111/j.1600-0587.2011.06999.x}.

\bibitem[Holbrook et~al.(2020)Holbrook, Lumley, and
  Gillen]{holbrook.estimating.2020}
A.~Holbrook, T.~Lumley, and D.~Gillen.
\newblock Estimating prediction error for complex samples.
\newblock \emph{Can. J. Stat.}, 48\penalty0 (2):\penalty0 204--221, 2020.
\newblock \doi{10.1002/cjs.11527}.

\bibitem[Horvitz and Thompson(1952)]{horvitz.generalization.1952}
D.~G. Horvitz and D.~J. Thompson.
\newblock A generalization of sampling without replacement from a finite
  universe.
\newblock \emph{J. Am. Stat. Assoc.}, 47\penalty0 (260):\penalty0 663--685,
  1952.
\newblock \doi{10.2307/2280784}.

\bibitem[Huang et~al.(2006)Huang, Zhu, and Siew]{Huang:2006}
G.-B. Huang, Q.-Y. Zhu, and C.-K. Siew.
\newblock Extreme learning machine: Theory and applications.
\newblock \emph{Neurocomputing}, 70\penalty0 (1-3):\penalty0 489--501, 2006.
\newblock \doi{10.1016/j.neucom.2005.12.126}.

\bibitem[Huang et~al.(2019)Huang, Chen, Liu, Chen, Huang, Liu, Zhao, Zhang, and
  Wang]{huang.hierarchical.2019}
W.~Huang, E.~Chen, Q.~Liu, Y.~Chen, Z.~Huang, Y.~Liu, Z.~Zhao, D.~Zhang, and
  S.~Wang.
\newblock Hierarchical multi-label text classification: An attention-based
  recurrent network approach.
\newblock In \emph{Proceedings of the 28th ACM International Conference on
  Information and Knowledge Management}, pages 1051--1060, 2019.

\bibitem[Hulten et~al.(2001)Hulten, Spencer, and Domingos]{Hulten:2001}
G.~Hulten, L.~Spencer, and P.~Domingos.
\newblock Mining time-changing data streams.
\newblock In \emph{Proceedings of the Seventh ACM SIGKDD International
  Conference on Knowledge Discovery and Data Mining}, pages 97--106, 2001.

\bibitem[Hyndman and Athanasopoulos(2018)]{Hyndman:2018}
R.~J. Hyndman and G.~Athanasopoulos.
\newblock \emph{Forecasting: Principles and Practice}.
\newblock OTexts, Melbourne, 2018.

\bibitem[Hájek(1971)]{Hajek1971}
J.~Hájek.
\newblock Comment on "an essay on the logical foundations of survey sampling".
\newblock In V.~Godambe and D.~Sprott, editors, \emph{Foundations of
  Statistical Inference}, page 236. Holt, Rinehart and Winston, 1971.

\bibitem[{Intergovernmental Panel on Climate Change}(2022)]{ipcc2022climate}
{Intergovernmental Panel on Climate Change}.
\newblock Climate change 2022: Impacts, adaptation and vulnerability.
\newblock Technical report, 2022.

\bibitem[{International Labour Organization}(2012)]{ISCO:2012}
{International Labour Organization}.
\newblock International standard classification of occupations: Isco-08.
\newblock Technical report, 2012.

\bibitem[James et~al.(2013)James, Witten, Hastie, and Tibshirani]{James:2013}
G.~James, D.~Witten, T.~Hastie, and R.~Tibshirani.
\newblock \emph{An Introduction to Statistical Learning}.
\newblock Springer, New York, 2013.
\newblock \doi{10.1007/978-1-4614-7138-7}.

\bibitem[Japkowicz and Shah(2011)]{japkowicz.evaluating.2011a}
N.~Japkowicz and M.~Shah.
\newblock \emph{Evaluating {{Learning Algorithms}}: {{A Classification
  Perspective}}}.
\newblock Cambridge University Press, Cambridge, 2011.
\newblock \doi{10.1017/CBO9780511921803}.

\bibitem[Kiritchenko et~al.(2005)Kiritchenko, Matwin, and
  Famili]{kiritchenko.functional.2005}
S.~Kiritchenko, S.~Matwin, and A.~F. Famili.
\newblock Functional annotation of genes using hierarchical text
  categorization.
\newblock In \emph{Proceedings of the BioLINK SIG: Linking Literature,
  Information and Knowledge for Biology (Held at ISMB-05)}, 2005.

\bibitem[Kohavi(1995)]{kohavi.study.1995}
R.~Kohavi.
\newblock A study of cross-validation and bootstrap for accuracy estimation and
  model selection.
\newblock In \emph{Proceedings of the 14th International Joint Conference on
  Artificial Intelligence}, pages 1137--1143, 1995.

\bibitem[Kosmopoulos et~al.(2015)Kosmopoulos, Partalas, Gaussier, Paliouras,
  and Androutsopoulos]{kosmopoulos.evaluation.2015}
A.~Kosmopoulos, I.~Partalas, E.~Gaussier, G.~Paliouras, and I.~Androutsopoulos.
\newblock Evaluation measures for hierarchical classification: A unified view
  and novel approaches.
\newblock \emph{Data Min. Knowl. Disc.}, 29\penalty0 (3):\penalty0 820--865,
  2015.
\newblock \doi{10.1007/s10618-014-0382-x}.

\bibitem[Kowsari et~al.(2017)Kowsari, Brown, Heidarysafa, Jafari~Meimandi,
  Gerber, and Barnes]{kowsari.hdltex.2017}
K.~Kowsari, D.~E. Brown, M.~Heidarysafa, K.~Jafari~Meimandi, M.~S. Gerber, and
  L.~E. Barnes.
\newblock Hdltex: Hierarchical deep learning for text classification.
\newblock In \emph{Proceedings of the 16th IEEE International Conference on
  Machine Learning and Applications (ICMLA)}, pages 364--371, 2017.

\bibitem[Kunjan et~al.(2021)Kunjan, Grummett, Pope, Powers, Fitzgibbon,
  Bastiampillai, Battersby, and Lewis]{Kunjan:2021}
S.~Kunjan, T.~S. Grummett, K.~J. Pope, D.~M. Powers, S.~P. Fitzgibbon,
  T.~Bastiampillai, M.~Battersby, and T.~W. Lewis.
\newblock The necessity of leave one subject out (loso) cross validation for
  eeg disease diagnosis.
\newblock In \emph{Proceedings of the 14th International Conference on Brain
  Informatics, Virtual Event}, pages 558--567, 2021.

\bibitem[K{\"u}nzel et~al.(2020)K{\"u}nzel, M{\"o}ller, Lindner, Goerdt, Nadal,
  Schmid, Schmitz-Valckenberg, Holz, Fleckenstein,
  et~al.]{kunzel2020determinants}
S.~H. K{\"u}nzel, P.~T. M{\"o}ller, M.~Lindner, L.~Goerdt, J.~Nadal, M.~Schmid,
  S.~Schmitz-Valckenberg, F.~G. Holz, M.~Fleckenstein, et~al.
\newblock Determinants of quality of life in geographic atrophy secondary to
  age-related macular degeneration.
\newblock \emph{Invest. Ophth. Vis. Sci.}, 61\penalty0 (5):\penalty0 63--63,
  2020.
\newblock \doi{10.1167/iovs.61.5.63}.

\bibitem[Le~Rest et~al.(2014)Le~Rest, Pinaud, Monestiez, Chadoeuf, and
  Bretagnolle]{Rest:2014}
K.~Le~Rest, D.~Pinaud, P.~Monestiez, J.~Chadoeuf, and V.~Bretagnolle.
\newblock Spatial leave-one-out cross-validation for variable selection in the
  presence of spatial autocorrelation.
\newblock \emph{Global Ecol. Biogeogr.}, 23\penalty0 (7):\penalty0 811--820,
  2014.
\newblock \doi{10.1111/geb.12161}.

\bibitem[Lu et~al.(2018)Lu, Liu, Dong, Gu, Gama, and Zhang]{Lu:2018}
J.~Lu, A.~Liu, F.~Dong, F.~Gu, J.~Gama, and G.~Zhang.
\newblock Learning under concept drift: A review.
\newblock \emph{IEEE T. Knowl. Data En.}, 31\penalty0 (12):\penalty0
  2346--2363, 2018.
\newblock \doi{10.1109/TKDE.2018.2876857}.

\bibitem[Mao et~al.(2019)Mao, Tian, Han, and Ren]{mao.hierarchical.2019}
Y.~Mao, J.~Tian, J.~Han, and X.~Ren.
\newblock Hierarchical text classification with reinforced label assignment.
\newblock In \emph{Proceedings of the 2019 Conference on Empirical Methods in
  Natural Language Processing and the 9th International Joint Conference on
  Natural Language Processing (EMNLP-IJCNLP)}, pages 445--455, 2019.

\bibitem[McCallum et~al.(1998)McCallum, Rosenfeld, Mitchell, and
  Ng]{mccallum.improving.1998}
A.~McCallum, R.~Rosenfeld, T.~M. Mitchell, and A.~Y. Ng.
\newblock Improving text classification by shrinkage in a hierarchy of classes.
\newblock In \emph{Proceedings of the Fifteenth International Conference on
  Machine Learning}, pages 359--367, 1998.

\bibitem[Meertens et~al.(2022)Meertens, Diks, van~den Herik, and
  Takes]{Meertens:2022}
Q.~Meertens, C.~Diks, H.~van~den Herik, and F.~Takes.
\newblock Improving the output quality of official statistics based on machine
  learning algorithms.
\newblock \emph{J. Off. Stat.}, 38\penalty0 (2):\penalty0 485--508, 2022.
\newblock \doi{10.2478/jos-2022-0023}.

\bibitem[Minku et~al.(2009)Minku, White, and Yao]{Minku:2009}
L.~L. Minku, A.~P. White, and X.~Yao.
\newblock The impact of diversity on online ensemble learning in the presence
  of concept drift.
\newblock \emph{IEEE T. Knowl. Data En.}, 22\penalty0 (5):\penalty0 730--742,
  2009.
\newblock \doi{10.1109/TKDE.2009.156}.

\bibitem[Molinaro et~al.(2005)Molinaro, Simon, and Pfeiffer]{Molinaro:2005}
A.~M. Molinaro, R.~Simon, and R.~M. Pfeiffer.
\newblock Prediction error estimation: a comparison of resampling methods.
\newblock \emph{Bioinformatics}, 21\penalty0 (15):\penalty0 3301--3307, 2005.
\newblock \doi{10.1093/bioinformatics/bti499}.

\bibitem[Peng et~al.(2018)Peng, Li, He, Liu, Bao, Wang, Song, and
  Yang]{peng.largescale.2018}
H.~Peng, J.~Li, Y.~He, Y.~Liu, M.~Bao, L.~Wang, Y.~Song, and Q.~Yang.
\newblock Large-scale hierarchical text classification with recursively
  regularized deep graph-cnn.
\newblock In \emph{Proceedings of the 2018 World Wide Web Conference}, pages
  1063--1072, 2018.

\bibitem[Pfau et~al.(2020)Pfau, von~der Emde, Dysli, M{\"o}ller, Thiele,
  Lindner, Schmid, Rubin, Fleckenstein, et~al.]{pfau2020determinants}
M.~Pfau, L.~von~der Emde, C.~Dysli, P.~T. M{\"o}ller, S.~Thiele, M.~Lindner,
  M.~Schmid, D.~L. Rubin, M.~Fleckenstein, et~al.
\newblock Determinants of cone and rod functions in geographic atrophy:
  {AI}-based structure-function correlation.
\newblock \emph{Am. J. Ophthalmol.}, 217:\penalty0 162--173, 2020.
\newblock \doi{10.1016/j.ajo.2020.04.003}.

\bibitem[Raschka(2020)]{raschka.model.2020a}
S.~Raschka.
\newblock Model evaluation, model selection, and algorithm selection in machine
  learning.
\newblock {arXiv:1811.12808}, 2020.

\bibitem[Roberts et~al.(2017)Roberts, Bahn, Ciuti, Boyce, Elith,
  {Guillera-Arroita}, Hauenstein, {Lahoz-Monfort}, Schr{\"o}der,
  et~al.]{Roberts:2017}
D.~R. Roberts, V.~Bahn, S.~Ciuti, M.~S. Boyce, J.~Elith, G.~{Guillera-Arroita},
  S.~Hauenstein, J.~J. {Lahoz-Monfort}, B.~Schr{\"o}der, et~al.
\newblock Cross-validation strategies for data with temporal, spatial,
  hierarchical, or phylogenetic structure.
\newblock \emph{Ecography}, 40\penalty0 (8):\penalty0 913--929, 2017.
\newblock \doi{10.1111/ecog.02881}.

\bibitem[Ru{\ss} and Brenning(2010)]{Russ:2010}
G.~Ru{\ss} and A.~Brenning.
\newblock Data mining in precision agriculture: Management of spatial
  information.
\newblock In \emph{Proceedings of the 13th International Conference on
  Information Processing and Management of Uncertainty in Knowledge-Based
  Systems}, pages 350--359, 2010.

\bibitem[Rutkowski et~al.(2014)Rutkowski, Jaworski, Pietruczuk, and
  Duda]{Rutkowski:2014}
L.~Rutkowski, M.~Jaworski, L.~Pietruczuk, and P.~Duda.
\newblock A new method for data stream mining based on the misclassification
  error.
\newblock \emph{IEEE T. Neur. Net. Lear.}, 26\penalty0 (5):\penalty0
  1048--1059, 2014.
\newblock \doi{10.1109/TNNLS.2014.2333557}.

\bibitem[Saeb et~al.(2017)Saeb, Lonini, Jayaraman, Mohr, and
  Kording]{Saeb:2017}
S.~Saeb, L.~Lonini, A.~Jayaraman, D.~C. Mohr, and K.~P. Kording.
\newblock The need to approximate the use-case in clinical machine learning.
\newblock \emph{Gigascience}, 6\penalty0 (5):\penalty0 gix019, 2017.
\newblock \doi{10.1093/gigascience/gix019}.

\bibitem[Sarndal(1980)]{sarndal.pinverse.1980}
C.~E. Sarndal.
\newblock On {$\pi$}-inverse weighting versus best linear unbiased weighting in
  probability sampling.
\newblock \emph{Biometrika}, 67\penalty0 (3):\penalty0 639--650, 1980.
\newblock \doi{10.2307/2335134}.

\bibitem[Schratz and Becker(2022)]{Schratz:2022}
P.~Schratz and M.~Becker.
\newblock \emph{mlr3spatiotempcv: Spatiotemporal resampling methods for
  'mlr3'}, 2022.
\newblock R package version 2.0.3.

\bibitem[Schratz et~al.(2019)Schratz, Muenchow, Iturritxa, Richter, and
  Brenning]{Schratz:2019}
P.~Schratz, J.~Muenchow, E.~Iturritxa, J.~Richter, and A.~Brenning.
\newblock Hyperparameter tuning and performance assessment of statistical and
  machine-learning algorithms using spatial data.
\newblock \emph{Ecol. Model.}, 406:\penalty0 109--120, 2019.
\newblock \doi{10.1016/j.ecolmodel.2019.06.002}.

\bibitem[Schratz et~al.(2021)Schratz, Becker, Lang, and Brenning]{Schratz:2021}
P.~Schratz, M.~Becker, M.~Lang, and A.~Brenning.
\newblock Mlr3spatiotempcv: {Spatiotemporal} resampling methods for machine
  learning in {R}.
\newblock {arXiv:2110.12674}, 2021.

\bibitem[Schreuder et~al.(2001)Schreuder, Gregoire, and
  Weyer]{schreuder.what.2001}
H.~T. Schreuder, T.~G. Gregoire, and J.~P. Weyer.
\newblock For what applications can probability and non-probability sampling be
  used?
\newblock \emph{Environ. Monit. Assess.}, 66:\penalty0 281--291, 2001.
\newblock \doi{10.1023/A:1006316418865}.

\bibitem[Simon(2007)]{simon.resampling.2007}
R.~Simon.
\newblock Resampling strategies for model assessment and selection.
\newblock In W.~Dubitzky, M.~Granzow, and D.~Berrar, editors,
  \emph{Fundamentals of Data Mining in Genomics and Proteomics}, pages
  173--186. Springer, Boston, MA, 2007.
\newblock \doi{10.1007/978-0-387-47509-7_8}.

\bibitem[Skinner and Wakefield(2017)]{skinner.introduction.2017}
C.~Skinner and J.~Wakefield.
\newblock Introduction to the design and analysis of complex survey data.
\newblock \emph{Stat. Sci.}, 32\penalty0 (2):\penalty0 165--175, 2017.
\newblock \doi{10.1214/17-STS614}.

\bibitem[Sokolova and Lapalme(2009)]{Sokolova:2009}
M.~Sokolova and G.~Lapalme.
\newblock A systematic analysis of performance measures for classification
  tasks.
\newblock \emph{Inform. Process. Manag.}, 45\penalty0 (4):\penalty0 427--437,
  2009.
\newblock \doi{10.1016/j.ipm.2009.03.002}.

\bibitem[Steyerberg(2019)]{Steyerberg:2019}
E.~W. Steyerberg.
\newblock \emph{Clinical Prediction Models: A Practical Approach to
  Development, Validation, and Updating}.
\newblock Springer, New York, 2 edition, 2019.

\bibitem[Sun and Lim(2001)]{sun.hierarchical.2001}
A.~Sun and E.-P. Lim.
\newblock Hierarchical text classification and evaluation.
\newblock In \emph{Proceedings of the 2001 IEEE International Conference on
  Data Mining}, pages 521--528, 2001.

\bibitem[Toth and Eltinge(2011)]{toth.building.2011a}
D.~Toth and J.~L. Eltinge.
\newblock Building consistent regression trees from complex sample data.
\newblock \emph{J. Am. Stat. Assoc.}, 106\penalty0 (496):\penalty0 1626--1636,
  2011.
\newblock \doi{10.1198/jasa.2011.tm10383}.

\bibitem[Tougui et~al.(2021)Tougui, Jilbab, and El~Mhamdi]{Tougui:2021}
I.~Tougui, A.~Jilbab, and J.~El~Mhamdi.
\newblock Impact of the choice of cross-validation techniques on the results of
  machine learning-based diagnostic applications.
\newblock \emph{Healthc. Inform. Res.}, 27\penalty0 (3):\penalty0 189--199,
  2021.
\newblock \doi{10.4258/hir.2021.27.3.189}.

\bibitem[{United Nations - Department of Economic and Social
  Affairs}(2022)]{un2022handbook}
{United Nations - Department of Economic and Social Affairs}.
\newblock The handbook on management and organization of national statistical
  systems.
\newblock Technical report, 2022.

\bibitem[{United Nations Economic Commission for
  Europe}(2011)]{unece2011impact}
{United Nations Economic Commission for Europe}.
\newblock The impact of globalization on national accounts.
\newblock Technical report, 2011.

\bibitem[Vach(2012)]{Vach:2012}
W.~Vach.
\newblock \emph{Regression Models as a Tool in Medical Research}.
\newblock CRC Press, Boca Raton, 2012.
\newblock \doi{10.1201/b12925}.

\bibitem[Valavi et~al.(2018)Valavi, Elith, {Lahoz-Monfort}, and
  {Guillera-Arroita}]{valavi.blockcv.2018}
R.~Valavi, J.~Elith, J.~J. {Lahoz-Monfort}, and G.~{Guillera-Arroita}.
\newblock blockcv: An r package for generating spatially or environmentally
  separated folds for k-fold cross-validation of species distribution models.
\newblock \emph{Methods Ecol. Evol.}, 10\penalty0 (2), 2018.
\newblock \doi{10.1111/2041-210X.13107}.

\bibitem[Valliant et~al.(2018)Valliant, Dever, and
  Kreuter]{valliant.practical.2018}
R.~Valliant, J.~A. Dever, and F.~Kreuter.
\newblock \emph{Practical Tools for Designing and Weighting Survey Samples}.
\newblock Statistics for Social and Behavioral Sciences. Springer, Cham, 2018.
\newblock \doi{10.1007/978-3-319-93632-1}.

\bibitem[Wang et~al.(1999)Wang, Zhou, and Liew]{wang.building.1999}
K.~Wang, S.~Zhou, and S.~C. Liew.
\newblock Building hierarchical classifiers using class proximity.
\newblock In \emph{Proceedings of the 25th International Conference on Very
  Large Data Bases}, page 363–374, 1999.

\bibitem[Webb et~al.(2016)Webb, Hyde, Cao, Nguyen, and Petitjean]{Webb:2016}
G.~I. Webb, R.~Hyde, H.~Cao, H.~L. Nguyen, and F.~Petitjean.
\newblock Characterizing concept drift.
\newblock \emph{Data Min. Knowl. Disc.}, 30\penalty0 (4):\penalty0 964--994,
  2016.
\newblock \doi{10.1007/s10618-015-0448-4}.

\bibitem[Wehrmann et~al.(2018)Wehrmann, Cerri, and
  Barros]{wehrmann.hierarchical.2018}
J.~Wehrmann, R.~Cerri, and R.~Barros.
\newblock Hierarchical multi-label classification networks.
\newblock In \emph{Proceedings of the 35th International Conference on Machine
  Learning}, pages 5075--5084, 2018.

\bibitem[Wenger and Olden(2012)]{Wenger:2012}
S.~J. Wenger and J.~D. Olden.
\newblock Assessing transferability of ecological models: An underappreciated
  aspect of statistical validation.
\newblock \emph{Methods Ecol. Evol.}, 3\penalty0 (2):\penalty0 260--267, 2012.
\newblock \doi{10.1111/j.2041-210X.2011.00170.x}.

\bibitem[{World Bank Group}(2019)]{worldbank2019changing}
{World Bank Group}.
\newblock \emph{The Changing Nature of Work}.
\newblock World Development Report. World Bank, 2019.

\bibitem[Wu et~al.(2019)Wu, Tygert, and LeCun]{wu.hierarchical.2019}
C.~Wu, M.~Tygert, and Y.~LeCun.
\newblock A hierarchical loss and its problems when classifying
  non-hierarchically.
\newblock \emph{PLOS ONE}, 14\penalty0 (12):\penalty0 e0226222, 2019.
\newblock \doi{10.1371/journal.pone.0226222}.

\bibitem[Xu and Wang(2017)]{Xu:2017}
S.~Xu and J.~Wang.
\newblock Dynamic extreme learning machine for data stream classification.
\newblock \emph{Neurocomputing}, 238:\penalty0 433--449, 2017.
\newblock \doi{10.1016/j.neucom.2016.12.078}.

\bibitem[Xue et~al.(2008)Xue, Xing, Yang, and Yu]{xue.deep.2008}
G.-R. Xue, D.~Xing, Q.~Yang, and Y.~Yu.
\newblock Deep classification in large-scale text hierarchies.
\newblock In \emph{Proceedings of the 31st Annual International {{ACM SIGIR}}
  Conference on Research and Development in Information Retrieval}, pages
  619--626, 2008.

\bibitem[Yung et~al.(2022)Yung, Tam, Buelens, Chipman, Dumpert, Ascari, Rocci,
  Burger, and Choi]{QF4SA:2022}
W.~Yung, S.-M. Tam, B.~Buelens, H.~Chipman, F.~Dumpert, G.~Ascari, F.~Rocci,
  J.~Burger, and I.~K. Choi.
\newblock A quality framework for statistical algorithms.
\newblock \emph{Stat. J. IAOS}, 38\penalty0 (1):\penalty0 291--308, 2022.
\newblock \doi{10.3233/SJI-210875}.

\bibitem[Zhang et~al.(2019)Zhang, Zhang, Shi, Almpanidis, Fan, and
  Shen]{Zhang:2019}
C.~Zhang, Y.~Zhang, X.~Shi, G.~Almpanidis, G.~Fan, and X.~Shen.
\newblock On incremental learning for gradient boosting decision trees.
\newblock \emph{Neural Process. Lett.}, 50:\penalty0 957--987, 2019.
\newblock \doi{10.1007/s11063-019-09999-3}.

\bibitem[Zhang et~al.(2017)Zhang, Chu, Li, Hu, and Wu]{Zhang:2017}
Y.~Zhang, G.~Chu, P.~Li, X.~Hu, and X.~Wu.
\newblock Three-layer concept drifting detection in text data streams.
\newblock \emph{Neurocomputing}, 260:\penalty0 393--403, 2017.
\newblock \doi{10.1016/j.neucom.2017.04.047}.

\bibitem[Zhou et~al.(2011)Zhou, Xiao, and Wu]{zhou.hierarchical.2011}
D.~Zhou, L.~Xiao, and M.~Wu.
\newblock Hierarchical classification via orthogonal transfer.
\newblock In \emph{Proceedings of the 28th International Conference on Machine
  Learning (ICML)}, pages 801--808, 2011.

\bibitem[{\v{Z}}liobait{\.e}(2014)]{Zliobaite:2014}
I.~{\v{Z}}liobait{\.e}.
\newblock Controlled permutations for testing adaptive learning models.
\newblock \emph{Knowl. Inf. Syst.}, 39\penalty0 (3):\penalty0 565--578, 2014.
\newblock \doi{10.1007/s10115-013-0629-7}.

\end{thebibliography}

\end{document}